\documentclass{article} %

\usepackage{xhfill}
\usepackage{array}
\usepackage{caption}
\usepackage{subcaption}
\usepackage[hidelinks]{hyperref}
\usepackage{url}
\usepackage{graphicx}
\usepackage{booktabs}
\usepackage{multirow}
\usepackage{wrapfig}
\usepackage{microtype}
\usepackage{titling}
\usepackage{glossaries-extra}
\usepackage{enumitem}
\usepackage{wrapfig}
\usepackage{bm}
\usepackage{amsmath}
\usepackage{amsfonts}
\usepackage[colorinlistoftodos]{todonotes}

\newcolumntype{M}[1]{>{\centering\arraybackslash}m{#1}}

\usepackage[capitalize,nameinlink]{cleveref}
\crefdefaultlabelformat{(#2\color{muted-indigo}\textbf{#1}#3)}

\DeclareCaptionFormat{captionsize}{\fontsize{8}{10}\selectfont#1#2#3}
\DeclareCaptionFormat{subcaptionsize}{\fontsize{7}{8}\selectfont#1#2#3}
\usepackage{pgfplots}
\usepackage{pgfplotstable}
\usepackage{tikz}
\usetikzlibrary{shapes.geometric}
\usetikzlibrary{patterns}
\usetikzlibrary{shadows.blur}
\usetikzlibrary{spy}
\usetikzlibrary{decorations.fractals,shapes.misc}

\definecolor{muted-indigo}{RGB}{51,34,136}
\definecolor{muted-cyan}{RGB}{136,204,238}
\definecolor{muted-teal}{RGB}{68,170,153}
\definecolor{muted-green}{RGB}{17,119,51}
\definecolor{muted-olive}{RGB}{153,153,51}
\definecolor{muted-sand}{RGB}{221,204,119}
\definecolor{muted-rose}{RGB}{221,102,119}
\definecolor{muted-wine}{RGB}{136,34,85}
\definecolor{muted-purple}{RGB}{170,68,153}

\definecolor{vibrant-blue}{RGB}{0,119,187}
\definecolor{vibrant-cyan}{RGB}{51,187,238}
\definecolor{vibrant-teal}{RGB}{0,153,136}
\definecolor{vibrant-orange}{RGB}{238,119,51}
\definecolor{vibrant-red}{RGB}{204,51,17}
\definecolor{vibrant-magenta}{RGB}{238,51,119}
\definecolor{vibrant-grey}{RGB}{187,187,187}

\tikzset{
  green-stimulus/.style={
    draw=muted-teal,
    fill=muted-teal!50,
    thick,
  },
  purple-stimulus/.style={
    draw=muted-purple,
    fill=muted-purple!50,
    thick,
  },
  diamond-stimulus/.style={
    diamond,
    minimum height=25pt,
    minimum width=25pt,
  },
  circle-stimulus/.style={
    circle,
    minimum height=25pt,
    minimum width=25pt,
  },
  null-prediction/.style={
    dashed,
    very thick,
    color=white!60!black
  },
  trend-line/.style={
    no marks,
    very thick,
    color=muted-wine,
  },
  pe-mark/.style={
    star,
    thick,
    draw=black,
    minimum width=1pt,
  },
  zs-interp-mark/.style={
    regular polygon,regular polygon sides=5,
    draw=black,
    thick,
    fill=vibrant-teal,
    minimum width=1pt,
    mark size=1pt,
  },
  fe-interp-mark/.style={
    star,star points=7,star point ratio=0.8,
    draw=black,
    thick,
    fill=vibrant-blue,
    minimum width=1pt,
  },
  corr-interp-mark/.style={
    rectangle,
    draw=black,
    thick,
    fill=vibrant-cyan,
    minimum width=1pt,
  },
  heatmap-spy/.style={
    spy scope={%
      magnification=10,
      width=50pt,
      height=18pt,
      connect spies,
      every spy on node/.style={
        rounded rectangle,
      },
      every spy in node/.style={
        draw,
        rounded rectangle,
      },
    },
  },
}

\pgfplotsset{
  comparison-axis/.style={
    width=80pt,
    height=65pt,
    symbolic x coords={``dax'',``fep''},
    xtick=data,
    axis y line=none,
    axis x line*=bottom,
    enlarge x limits=0.5,
    ticklabel style={font=\tiny}
  },
  comparison-plot/.style={
    ybar,
    thick,
    muted-cyan,
    fill=muted-cyan!50,
  },
  points-in-a-plane-axis/.style={
    width=108pt,
    height=108pt,
    xmin=-7, xmax=7,
    ymin=-7, ymax=7,
    enlargelimits=false,
    thick,
    axis y line=left,
    axis x line=bottom,
    ticklabel style={font=\fontsize{2.5}{4}},
    xticklabel style={yshift=3pt},
    yticklabel style={xshift=3pt},
  },
  points-in-a-plane/.style={
    only marks,
    mark size=0.5pt,
    samples=200,
  },
  effect-axis/.style={
    yticklabel style={font=\fontsize{1}{2}},
    ylabel={$\operatorname{EvR}(\family)$},
    ylabel style={yshift=-12pt},
    label style={font=\scriptsize},
    enlarge y limits=0.1,
  },
  effect-scatter/.style={
    effect-axis,
    xticklabel style={font=\fontsize{1}{2}},
    ylabel={$\operatorname{EvR}(\family)$},
    xlabel={$\operatorname{FLB}(\family)$},
    ymin=-0.5, ymax=2.5,
    xmin=-2.5, xmax=2.5,
    xtick distance=1,
    ytick distance=1,
    scale only axis,
    clip=false,
    xlabel style={yshift=2pt},
    ytick style={draw=none},
    axis x line=bottom,
    axis y line=left,
    ylabel style={yshift=-6pt},
    xlabel style={yshift=2pt},
  },
  effect-bar/.style={
    effect-axis,
    clip=false,
    ymin=0, ymax=0.5,
    major x tick style=transparent,
    xlabel style={yshift=-22pt},
    xticklabel style={
      rotate=45,
    },
    ybar=2*\pgflinewidth,
    bar width=8pt,
    xtick=data,
    scaled y ticks=false,
    legend cell align=left,
    legend style={
      at={(1,1.05)},
      anchor=south east,
      column sep=1ex,
    },
  },
  effect-scatter-plot/.style={
    only marks,
    scatter,
    scatter/@pre marker code/.append style={/tikz/mark size=1pt},
    error bars/.cd,
    y dir=both,
    y explicit,
    x dir=both,
    x explicit,
    error bar style={
      draw=black,
      thin,
    },
    error mark options={
      draw=black,
      thin,
      mark size=1pt,
      rotate=90,
    },
  },
  effect-bar-plot/.style={
    error bars/.cd,
    y dir=both,
    y explicit,
  },
  heatmap-axis/.style={
    xmin=-0.04, xmax=1.04,
    ymin=-0.04, ymax=1.04,
    ytick distance=1.0,
    xtick distance=0.5,
    grid=none,
    clip=false,
    ticklabel style={font=\fontsize{2.5}{4}},
    axis line style={draw=none},
    tick style={draw=none},
    axis y line*=left,
    axis x line*=bottom,
    xlabel near ticks,
    ylabel near ticks,
    xlabel={\tiny$\mathbf\propone$},
    xlabel style={yshift=-2pt},
    ylabel={\tiny$\mathbf\propzero$},
    ylabel style={rotate=-90,xshift=6pt},
  },
  interpolation-axis/.style={
    clip=false,
    width=108pt,
    height=108pt,
    ymin=0.0, ymax=0.5,
    xmin=0.0, xmax=0.25,
    enlarge y limits={abs=5pt},
    xtick distance=0.125,
    grid=major,
    ticklabel style={font=\fontsize{2.5}{4}},
    ylabel={$[\operatorname{ZS} - \text{\,equi-}\rho]$},
    ylabel style={font=\scriptsize,yshift=-16pt},
    xlabel={\textbf{interpolation distance} \\ {\tiny$\big(\sqrt{(0.5 - \propzero)^2 + \propone^2}\big)$}},
    xlabel style={font=\scriptsize,align=center},
  },
  interp-plot/.style={
    error bars/.cd,
    y dir=both,
    y fixed=0.1,
    error bar style={color=black, mark size=.4pt},
  },
}

\tikzset{
  declare function={invgaussx(\r,\s,\mu,\sigma)=\sigma * sqrt(-2*ln(\r))*cos(deg(2*pi*\s)) + \mu;},
  declare function={invgaussy(\r,\s,\mu,\sigma)=\sigma * sqrt(-2*ln(\r))*sin(deg(2*pi*\s)) + \mu;},
}

\setabbreviationstyle[long]{long}             %
\setabbreviationstyle[long-noshort]{long-em}  %

\newabbreviation[category=long-short]{ml}{ML}{machine learning}
\newabbreviation[category=long-noshort]{fc}{}{feature co-occurrence}

\newabbreviation[category=long-short]{flb}{$\operatorname{FLB}$}{feature-level bias}
\newabbreviation[category=long-short]{evr}{$\operatorname{EvR}$}{exemplar vs. rule propensity}

\newabbreviation[category=long-noshort]{zs}{}{zero shot}
\newabbreviation[category=long-noshort]{pe}{}{partial exposure}
\newabbreviation[category=long-noshort]{cc}{}{cue conflict}
\newabbreviation[category=long-noshort]{zsc}{}{zero-shot condition}
\newabbreviation[category=long-noshort]{pec}{}{partial-exposure condition}
\newabbreviation[category=long-noshort]{ccc}{}{cue-conflict condition}

\newabbreviation[category=long-short]{erl}{ER level}{exemplar-rule level}
\newabbreviation[category=long-short]{ccctrl}{CC control}{cue-conflict control}
\newabbreviation[category=long-short]{zsctrl}{ZS control}{zero-shot control}
\newabbreviation[category=long-short]{rnn}{RNN}{recurrent neural network}
\newabbreviation[category=long-short]{ood}{OOD}{out-of-domain}
\newabbreviation[category=long-short]{erm}{ERM}{empirical risk minimization}
\newabbreviation[category=long-short]{ntk}{NTK}{neural tangent kernel}
\newabbreviation[category=long-short]{nn}{NN}{neural network}
\newabbreviation[category=long-short]{icp}{ICP}{invariant causal prediction}
\newabbreviation[category=long-short]{resnet}{ResNet}{residual neural network}
\newabbreviation[category=long-short]{lstm}{LSTM}{long short-term memory}
\newabbreviation[category=long-short]{gnn}{GNN}{graph neural network}
\newabbreviation[category=long-short]{gp}{GP}{Gaussian process}
\newabbreviation[category=long-short]{rbf}{RBF}{radial basis function}
\newabbreviation[category=long-short]{glm}{GLM}{generalized linear model}
\newabbreviation[category=long-short]{relu}{ReLU}{rectified linear unit}
\newabbreviation[category=long-short]{celeba}{CelebA}{CelebFaces Attributes}
\newabbreviation[category=long-short]{imdb}{IMDb}{Internet Movie Database Movie Reviews}

\newcommand{\attribute}{\mathbf{z}}
\newcommand{\inputs}{\mathbf{x}}
\newcommand{\disc}{\attribute_\text{disc}}
\newcommand{\dist}{\attribute_\text{dist}}

\newcommand{\prop}{\pi}

\newcommand{\propzero}{\prop_0}
\newcommand{\propone}{\prop_1}
\newcommand{\corr}{\rho}

\newcommand{\propfe}{\prop^{\, \text{FE}}}
\newcommand{\propzs}{\prop^{\, \text{ZS}}\left(\prop^{\, \text{FE}}\right)}
\newcommand{\propeq}{\prop^{\, \text{EQ}}\left(\prop^{\, \text{FE}}\right)}

\newcommand{\family}{\mathcal{F}}
\newcommand{\model}{\hat{f}}
\newcommand{\fcc}{\model^{\, \text{CC}}}
\newcommand{\fzs}{\model^{\, \text{ZS}}}
\newcommand{\fpe}{\model^{\, \text{PE}}}

\glsdisablehyper

\usepgfplotslibrary{external}

\makeatletter
\def\blfootnote{\xdef\@thefnmark{}\@footnotetext}
\makeatother

\usepackage[accepted]{front-matter/icml-style/icml2022}

\begin{document}

\icmltitlerunning{Distinguishing rule- and exemplar-based generalization in learning systems}
\twocolumn[
\icmltitle{Distinguishing rule- and exemplar-based generalization \\
in learning systems}

\icmlsetsymbol{equal}{*}
\icmlsetsymbol{dm}{@}

\begin{icmlauthorlist}
\icmlauthor{Ishita Dasgupta}{equal,princeton,dm}
\icmlauthor{Erin Grant}{equal,ucb}
\icmlauthor{Thomas L.~Griffiths}{princeton}
\end{icmlauthorlist}

\icmlaffiliation{ucb}{Department of Electrical Engineering \& Computer Sciences, UC Berkeley}
\icmlaffiliation{princeton}{Departments of Psychology \& Computer Science, Princeton University}

\icmlcorrespondingauthor{Ishita Dasgupta}{\texttt{idg@deepmind.com}}
\icmlcorrespondingauthor{Erin Grant}{\texttt{eringrant@berkeley.edu}}

\icmlkeywords{cognitive psychology, representation learning}

\vskip 0.3in
]

\printAffiliationsAndNotice{\icmlEqualContribution}

\newcommand{\etal}{\textit{et~al}.~\ }
\newcommand{\eg}{\textit{e.g.,}~}
\newcommand{\ie}{\textit{i.e.,}~}
\newcommand{\viz}{\textit{viz.}~}
\newcommand{\cf}{\textit{c.f.,}~}

\begin{abstract}
Machine learning systems often do not share the same inductive biases as humans and, as a result, extrapolate or generalize in ways that are inconsistent with our expectations.
The trade-off between exemplar- and rule-based generalization has been studied extensively in cognitive psychology; in this work, we present a protocol inspired by these experimental approaches to probe the inductive biases that control this trade-off in category-learning systems.
We isolate two such inductive biases: feature-level bias (differences in which features are more readily learned) and exemplar or rule bias (differences in how these learned features are used for generalization).
We find that standard neural network models are feature-biased and exemplar-based, and discuss the implications of these findings for machine learning research on systematic generalization, fairness, and data augmentation.
\end{abstract}

\section{Introduction}
\label{sec:intro}

Extrapolation or generalization---decisions on unseen datapoints---is always underdetermined by data;
which particular extrapolation behavior a \gls{ml} system exhibits is determined by its inductive biases~\citep{mitchell1980need}. 
When those inductive biases are opaque---as is often the case with many modern \gls{ml} systems~\citep{geirhos2020shortcut,d2020underspecification}---we can instead turn to empirical investigation of the \emph{behavior} of a system to reveal the system's \emph{implicit} inductive biases.
Cognitive psychology provides a rich basis for experimental designs to study the often-opaque human cognitive system via its external behavior;
these designs can be leveraged to distinguish between competing hypotheses about a machine learning system's inductive biases as well~\citep[\eg][]{ritter2017cognitive,lake2018building,dasgupta2019analyzing}.

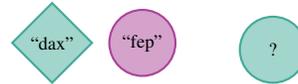
\begin{figure}[t]
  \footnotesize
  \centering

  \begin{tabular}{c    c}

    training examples & 
    extrapolation
    \\\midrule

    \begin{tikzpicture}
      \scriptsize
      \node[diamond-stimulus,green-stimulus] (d) at (0,0) {``dax''};
      \node[circle-stimulus,purple-stimulus] (d) at (1.2,0) {``fep''};
    \end{tikzpicture}
    &
    \begin{tikzpicture}
      \scriptsize
      \node[circle-stimulus,green-stimulus] (d) at (1,0) {?};
    \end{tikzpicture}

  \end{tabular}

  \caption{\textbf{Example of a data condition:} 
 Data often underdetermines a decision boundary; here, it is unclear whether shape or color determines object label (``dax'' vs ``fep''). How a learner extrapolates to new stimuli reveals inductive bias.
  }
  \label[figure]{fig:example}
  \vspace{-10pt}
\end{figure}

We draw on cognitive psychology to construct a protocol that isolates the inductive biases determining how an \gls{ml} system generalizes feature-based categories such as those in \cref{fig:example}.
A key property of such categorization problems is the presence of a \emph{distractor} dimension 
that does not play a causal role in the underlying category boundary;
the ground truth categorization is determined by a \emph{discriminant} dimension.
Such problems are ubiquitous in machine learning applications~\citep[\eg][]{beery2018recognition}, 
where learned associations between the distractor and the categorization label are termed 
``spurious''~\citep{arjovsky2019invariant}. 
The tendency to acquire (potentially harmful) spurious associations is an example of a downstream consequence of implicit inductive bias, and so characterizing such implicit inductive biases is of both theoretical and practical interest.

We use abstract problem settings such as that in \cref{fig:example} to identify and isolate two distinct inductive biases underlying feature-based category learning.
The first, \emph{feature-level bias}, expresses a preference for some features over others to support a decision boundary (\eg preferring shape over color).
The second, \emph{exemplar bias}---vs. \emph{rule bias}---expresses a preference for feature-dense (vs. feature-sparse) decision boundaries (\eg a boundary informed by both shape and color, vs. only one of the two features). Our protocol presents data conditions that manipulate \gls{fc}s observed during training such that the resulting extrapolation behavior is diagnostic of these inductive biases in the learner.
\blfootnote{
\hskip -1em
Code at \url{https://github.com/eringrant/icml-2022-rules-vs-exemplars}.}

The experimental setup underlying our training and testing conditions is similar to existing works in 
``combinatorial generalization''~\cite{andreas2016neural,johnson2017clevr}
and ``subgroup fairness''~\cite{sagawa2020distributionally,sagawa2020investigation}.
Our work also makes several independent contributions:
We identify and isolate two distinct inductive biases that affect extrapolation of feature-based categories, and
we examine these across models in an expository points-in-a-plane setting, as well as in more naturalistic text and image domains.
We demonstrate that existing measures of \gls{fc} and extrapolation behavior ~\citep[``spurious correlation'' and ``worst-group accuracy,''][]{sagawa2020investigation} are insufficient to characterize these inductive biases.
Finally, we consider the normative question: \emph{What extrapolation behavior is desirable for a given application}?
We provide a preliminary answer by discussing the relevance of the inductive biases we identify to related work in systematic generalization, fairness, and data augmentation.

\bgroup
\newcolumntype{M}[1]{>{\centering\arraybackslash}m{#1}}

\begin{figure*}[t]
  \footnotesize
  \centering

  \begin{tabular}{
      m{35pt}
      @{\hskip 10pt}
      M{100pt}
      M{50pt}
      @{\hskip 10pt}
      M{48pt}
      M{48pt}
      M{48pt}
    }

    \toprule
    &
    \multicolumn{2}{c}{observations} & 
    \multicolumn{3}{c}{ratio of predictions}
    \\
    \cmidrule(lr){2-3}
    \cmidrule(lrl){4-6}

    condition & 
    training examples & 
    extrapolation & 
    humans & 
    rule-based & 
    \mbox{\hspace{-6pt}exemplar-based}\\

    & 
    & 
    & 
    \scriptsize (\textit{shape-biased}) & 
    \scriptsize (\textit{no feature bias}) & 
    \scriptsize \mbox{\hspace{-6pt}(\textit{no feature bias})}
    \\\midrule

    \textbf{cue \linebreak conflict} &
    \begin{tikzpicture}
      \scriptsize
      \node[diamond-stimulus,green-stimulus] (d) at (0,0) {``dax''};
      \node[circle-stimulus,purple-stimulus] (d) at (1.2,0) {``fep''};
    \end{tikzpicture}
    &
    \begin{tikzpicture}
      \scriptsize
      \node[circle-stimulus,green-stimulus] (d) at (1,0) {?};
    \end{tikzpicture}
    &
    \begin{tikzpicture}[baseline]
      \begin{axis}[comparison-axis]
        \addplot[comparison-plot] coordinates {(``dax'',0) (``fep'',1)};
      \end{axis}
    \end{tikzpicture}
    &
    \begin{tikzpicture}[baseline]
      \begin{axis}[comparison-axis]
        \addplot[comparison-plot] coordinates {(``dax'',0.5) (``fep'',0.5)};
      \end{axis}
    \end{tikzpicture}
    &
    \begin{tikzpicture}[baseline]
      \begin{axis}[comparison-axis]
        \addplot[comparison-plot] coordinates {(``dax'',0.5) (``fep'',0.5)};
      \end{axis}
    \end{tikzpicture}
    \\

    \textbf{zero \linebreak shot} &
    \begin{tikzpicture}
      \scriptsize
      \node[diamond-stimulus,green-stimulus] (d) at (0,0) {``dax''};
      \node[diamond-stimulus,purple-stimulus] (d) at (1.2,0) {``fep''};
    \end{tikzpicture}
    &
    \begin{tikzpicture}
      \scriptsize
      \node[circle-stimulus,green-stimulus] (d) at (1,0) {?};
    \end{tikzpicture}
    &
    \begin{tikzpicture}[baseline]
      \begin{axis}[comparison-axis]
        \addplot[comparison-plot] coordinates {(``dax'',1) (``fep'',0)};
      \end{axis}
    \end{tikzpicture}
    &
    \begin{tikzpicture}[baseline]
      \begin{axis}[comparison-axis]
        \addplot[comparison-plot] coordinates {(``dax'',1) (``fep'',0)};
      \end{axis}
    \end{tikzpicture}
    &
    \begin{tikzpicture}[baseline]
      \begin{axis}[comparison-axis]
        \addplot[comparison-plot] coordinates {(``dax'',1) (``fep'',0)};
      \end{axis}
    \end{tikzpicture}
    \\

    \textbf{partial \linebreak exposure} &
    \begin{tikzpicture}
      \scriptsize
      \node[diamond-stimulus,green-stimulus] (d) at (-1.2,0) {``dax''};
      \node[circle-stimulus,purple-stimulus] (d) at (0,0) {``fep''};
      \node[diamond-stimulus,purple-stimulus] (d) at (1.2,0) {``fep''};
    \end{tikzpicture}
    &
    \begin{tikzpicture}
      \scriptsize
      \node[circle-stimulus,green-stimulus] (d) at (1,0) {?};
    \end{tikzpicture}
    &
    \begin{tikzpicture}[baseline]
      \begin{axis}[comparison-axis]
        \addplot[comparison-plot] coordinates {(``dax'',1) (``fep'',0)};
      \end{axis}
    \end{tikzpicture}
    &
    \begin{tikzpicture}[baseline]
      \begin{axis}[comparison-axis]
        \addplot[comparison-plot] coordinates {(``dax'',1) (``fep'',0)};
      \end{axis}
    \end{tikzpicture}
    &
    \begin{tikzpicture}[baseline]
      \begin{axis}[comparison-axis]
        \addplot[comparison-plot] coordinates {(``dax'',0.5) (``fep'',0.5)};
      \end{axis}
    \end{tikzpicture}
    \\\bottomrule

  \end{tabular}

  \caption{\textbf{Illustrative category learning experiment:} Training examples from the 3 independent training conditions, the extrapolation test, and characteristic behavior for learners with different inductive biases. 
  We formalize the training conditions in \cref{fig:setup-figure}.
  }
  \label[figure]{fig:expository}
\end{figure*}

\egroup
\section{Inductive biases in category learning}
\label{sec:expository}

We start by introducing the two inductive biases of interest.
\textbf{Feature-level bias} characterizes \emph{which} feature a system finds \textit{easier} or \textit{harder} to learn and thus which feature a system will utilize when both are associated with the category label.
This kind of feature-level bias has been studied extensively in human cognition~\citep{landau1988importance,hudson2005regularizing}, and specific feature-level biases---mostly notably the ``shape-bias,'' the tendency to generalize image category labels according to shape rather than according to color or texture---have been revisited in the context of recent neural network models~\citep{ritter17,hermann2019origins, geirhos2018imagenet}.
We examine feature-level bias for arbitrary features, as well as demonstrate how this bias interacts with---but is distinct from---another kind of inductive bias, to be discussed next.

\textbf{Exemplar} (or \textbf{rule}) \textbf{bias} characterizes \textit{how} a system uses features to inform decisions by trading off \emph{exemplar- and rule-based generalization}.
A rule-based decision is made on the basis of minimal features that support the category boundary~\citep[\eg][]{ashby1986varieties},
while an exemplar-based decision-maker generalizes on the basis of similarity to category exemplars~\citep[\eg][]{shepard1963stimulus},
invoking many or all features that underlie a category.
Extensive empirical work in cognitive psychology has found evidence of both kinds of generalization in humans~\citep{nosofsky1989rules,rips1989similarity,allen1991specializing,smith1994similarity}. 
This trade-off can be understood as a continuum that varies the number of features employed to discriminate between categories~\citep{pothos2005rules}.

Feature-level bias and exemplar bias are \textbf{practically relevant} because they describe how a learning system uses features to extrapolate, and different problem settings call for different ways of doing so.
An exemplar-based system that depends on all features, and is not invariant to any of them, suffers when not all feature combinations are observed and systematic generalization to unobserved combinations is expected~\citep{lake2018building, marcus2018deep, arjovsky2019invariant}.
On the other hand, a rule-based system that applies the same category decision rules across all data regions might over-generalize, which is undesirable in naturally occurring long-tailed distributions~\citep{feldman2020neural,feldman2020does,brown2020memorization}.
Diagnosing exemplar vs. rule bias is therefore of both theoretical and practical interest.  
In \cref{sec:celeba}, we give a concrete example in a fairness setting---where certain regions of the data support is underrepresented but we want comparable accuracy on these regions nonetheless---in which understanding the inductive biases of the learning system allows for a data intervention that improves performance.

We now build intuitions for how \textbf{the category learning paradigm in \cref{fig:expository} isolates feature-level bias and exemplar bias}.
The stimuli \cref{fig:expository} vary along two feature dimensions, shape and color.
Color determines the label of an object (\ie green objects are ``dax''; purple are ``fep'', using arbitrary names to emphasize that the category is novel to humans as well as to \gls{ml} systems).
Shape is unrelated to the underlying category structure and acts as a distractor.
Participants (either humans or artificial learning systems) are independently placed in three different conditions---\textbf{\gls{cc}}, \textbf{\gls{zs}}, or \textbf{\gls{pe}}---that vary in coverage of the feature space.
After observing the \textit{training examples}, the participant is presented with an \textit{extrapolation} test consisting of an example outside the support of feature combinations observed during training (\ie they must classify the green circle as a ``dax'' or a ``fep'').
We explain below how differences in classification behavior on this extrapolation test isolate feature-level bias as well as exemplar-vs-rule bias, but first: We encourage the reader to try the experiment themselves to examine their intuitions.

\textbf{\Gls{cc} \normalfont{(CC, top row, \cref{fig:expository})}.}
The data presented in this condition confound color and shape (\ie color and shape are equally predictive of the category boundary). How a system generalizes here directly measures its feature-level bias towards color or shape.

\emph{Characteristic behavior} (right half of \cref{fig:expository}).
Since humans have an established shape bias~\citep{landau1988importance}, we expect that humans will classify the test item according to the object that shares its shape, not its color; in this case,  as a ``fep.''
However, this inductive bias is independent of whether a reasoner is rule- or exemplar-based; neither  has an \textit{a priori} propensity for features, both are equally likely to classify the test item as a ``dax'' or a ``fep.''

\textbf{\Gls{zs} \normalfont{(ZS, middle row, \cref{fig:expository})}.}
This condition requires extrapolation to a new feature value ``zero-shot'' (\ie without prior exposure).
This setting is often used to examine \gls{ood} and compositional generalization in machine learning~\citep{xian2018zero}.
Behavior in this condition reveals whether the model has learned the discriminating features and whether it can extrapolate to new feature values, and thus acts as a baseline.

\emph{Characteristic behavior} (right half, \cref{fig:expository}).
Rule- and exemplar-based behavior in this condition is confounded.
A rule-based learner infers the minimal rule that color determines label, does not assign any predictive value to shape, and therefore classifies the extrapolation stimulus based on color as a ``dax.''
An exemplar-based learner categorizes based on the similarity along all feature dimensions of the extrapolation stimulus to category exemplars.
Neither training exemplars have any overlap with the test stimulus along the shape dimension, but the ``dax'' overlaps along the color dimension, and the learner categorizes it as a ``dax.''

\textbf{\Gls{pe} \normalfont{(PE, bottom row, \cref{fig:expository})}.}
  Compared to zero shot, participants in this condition also receive ``partial exposure'' to a new feature value (\ie \textit{circle}) along the shape dimension.
The extrapolation test in this condition is most similar to \textit{combinatorial zero-shot generalization}~\citep[\eg][]{lake2018generalization}, where the learner is exposed independently to all feature values but has to generalize to a new combination.

\emph{Characteristic behavior} (right half of \cref{fig:expository}).
This setting meaningfully distinguishes rule- and exemplar-based generalization. To understand this distinction,  we contrast this condition to the \gls{ccc}.
The addition of the purple diamond-shaped ``fep'' means the learner has seen both a diamond and a circle labeled ``fep''. A rule-based system takes this as direct evidence that shape is \textit{not} predictive of category label and classifies the extrapolation stimulus on the basis of color as a ``dax.'' This is typically also how humans extrapolate. This additional training example, however, does not impact an exemplar-based system, since it does not share any features with the extrapolation stimulus. The exemplar-based reasoner classifies on the basis of feature-overlap with training exemplars and is therefore indifferent, exactly as in the \gls{ccc}.

\begin{figure*}[t]
\centering
\begin{minipage}{.8\textwidth}
  \renewcommand{\arraystretch}{1.1}
  \footnotesize
  \centering

 \begin{tabular}{
      M{71pt}
      M{66pt}
      M{66pt}
      M{66pt}
      M{66pt}
    }

    \toprule

    &
    \multicolumn{3}{c}{training condition} & 
    \\
    \cmidrule(rl){2-4}
    feature space & 
    \gls{cc} & 
    \gls{zs} & 
    \gls{pe} &
    extrapolation
    \\

    \scriptsize ${\color{muted-purple}\propzero=0.5}$, ${\color{muted-teal}\propone=0.5}$
    & \scriptsize ${\color{muted-purple}\propzero=1.0}$, ${\color{muted-teal}\propone=0.0}$
    & \scriptsize ${\color{muted-purple}\propzero=0.0}$, ${\color{muted-teal}\propone=0.0}$
    & \scriptsize ${\color{muted-purple}\propzero=0.5}$, ${\color{muted-teal}\propone=0.0}$
    & \scriptsize \phantom{$\propzero=1.0$, }${\color{muted-teal}\propone=1.0}$
    \\\midrule

    \hspace{-10pt}
    \begin{tikzpicture}
      \scriptsize
      \fill[muted-purple!15] (0,0) -- (0,2.2) -- (1.1,2.2) -- (1.1,0);
      \fill[muted-teal!15] (1.1,0) -- (1.1,2.2) -- (2.2,2.2) -- (2.2,0);

      \draw[->,thick] (0, 0) -- (2.2, 0) node[midway,below,align=center] {\textbf{discriminant} (color)};
      \draw[->,thick] (0, 0) -- (0, 2.2) node[midway,left,rotate=90,anchor=south,align=center] {\textbf{distractor} (shape)};

      \node[diamond-stimulus,purple-stimulus] (off-off) at (0.6,0.6) {};
      \node[diamond-stimulus,green-stimulus] (on-off) at (1.6,0.6) {};
      \node[circle-stimulus,purple-stimulus] (off-on) at (0.6,1.6) {};
      \node[circle-stimulus,green-stimulus] (on-on) at (1.6,1.6) {};
    \end{tikzpicture}
    &
    \begin{tikzpicture}[baseline={(0,-0.35)}]
      \scriptsize
      \fill[muted-purple!15] (0,0) -- (0,2.2) -- (1.1,2.2) -- (1.1,0);
      \fill[muted-teal!15] (1.1,0) -- (1.1,2.2) -- (2.2,2.2) -- (2.2,0);

      \draw[->,thick] (0, 0) -- (2.2, 0);
      \draw[->,thick] (0, 0) -- (0, 2.2);

      \node[diamond-stimulus,green-stimulus] (on-off) at (1.6,0.6) {};
      \node[circle-stimulus,purple-stimulus] (off-on) at (0.6,1.6) {};
    \end{tikzpicture}
    &
    \begin{tikzpicture}[baseline={(0,-0.35)}]
      \scriptsize
      \fill[muted-purple!15] (0,0) -- (0,2.2) -- (1.1,2.2) -- (1.1,0);
      \fill[muted-teal!15] (1.1,0) -- (1.1,2.2) -- (2.2,2.2) -- (2.2,0);

      \draw[->,thick] (0, 0) -- (2.2, 0);
      \draw[->,thick] (0, 0) -- (0, 2.2);

      \node[diamond-stimulus,purple-stimulus] (off-off) at (0.6,0.6) {};
      \node[diamond-stimulus,green-stimulus] (on-off) at (1.6,0.6) {};
    \end{tikzpicture}
    &
    \begin{tikzpicture}[baseline={(0,-0.35)}]
      \scriptsize
      \fill[muted-purple!15] (0,0) -- (0,2.2) -- (1.1,2.2) -- (1.1,0);
      \fill[muted-teal!15] (1.1,0) -- (1.1,2.2) -- (2.2,2.2) -- (2.2,0);

      \draw[->,thick] (0, 0) -- (2.2, 0);
      \draw[->,thick] (0, 0) -- (0, 2.2);

      \node[diamond-stimulus,purple-stimulus] (off-off) at (0.6,0.6) {};
      \node[diamond-stimulus,green-stimulus] (on-off) at (1.6,0.6) {};
      \node[circle-stimulus,purple-stimulus] (off-on) at (0.6,1.6) {};
    \end{tikzpicture}
    & 
    \begin{tikzpicture}
      \scriptsize
      \fill[muted-purple!15] (0,0) -- (0,2.2) -- (1.1,2.2) -- (1.1,0);
      \fill[muted-teal!15] (1.1,0) -- (1.1,2.2) -- (2.2,2.2) -- (2.2,0);

      \draw[->,thick] (0, 0) -- (2.2, 0);
      \draw[->,thick] (0, 0) -- (0, 2.2);

      \node[circle-stimulus,green-stimulus] (on-on) at (1.6,1.6) {};
    \end{tikzpicture}

  \end{tabular}       
                     
  \caption{
    \textbf{Formalizing the illustrative experiment:} The experiment from \cref{fig:expository} expressed in terms of the formalism in \cref{sec:setup} with color as $\dist$ and shape as $\disc$. Background colors indicate the true category.
  }
  \label[figure]{fig:setup-figure}

\end{minipage}
\hfill
\begin{minipage}{.17\textwidth}
  \vspace{30pt}
    \begin{tikzpicture}
      \begin{axis}[
        heatmap-axis,
        width=110pt,
        height=110pt,
    ]
        \addplot graphics
        [xmin=-0.04, xmax=1.04, ymin=-0.04, ymax=1.04]
        {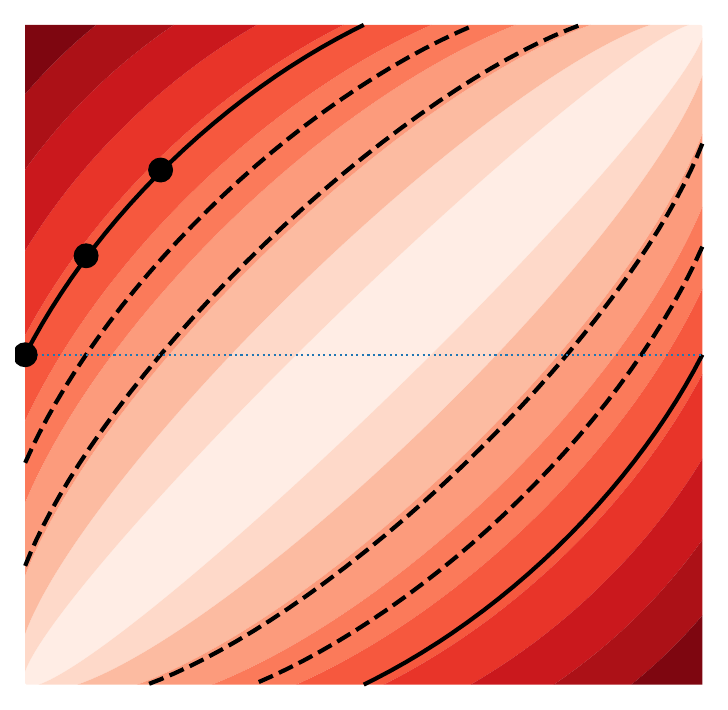};
       \end{axis}
     \end{tikzpicture}
    \caption{
      \textbf{Spurious correlation} (\cref{eq:spurious_corr}).
    }
    \label{fig:interpolation-heatmap}
\end{minipage}
\end{figure*}
\section{A protocol for measuring inductive bias}
\label{sec:setup}

We embed the structure of the category learning problem discussed in \cref{sec:expository} into a statistical learning problem that can be applied across domains to test black-box learners.

\textbf{Problem setting.}
We consider the \emph{oracle} compositional setting of \citet{andreas2019measuring} in which inputs are a composition of categorical attributes 
with two latent binary features, $\disc, \dist, \in \{0, 1\}$ that jointly determine the observation $\inputs \in \mathcal{X} \subset \mathbb{R}^d$ via some mapping $g : \{0,1\}^2 \to \mathcal{X}$;
see \cref{fig:setup-figure}. 
We consider the binary classification task of fitting a model $\hat{f} : \mathcal{X} \to \{0, 1\}$ from a given model family $\family$
to predict a class for each observation.
One of the underlying features, the \emph{discriminant}, $\disc$, defines the decision boundary; the other one, the \emph{distractor}, $\dist$, is not independently predictive of the label.

This specifies a generative process  $\inputs, \disc, \dist \sim p(\inputs \mid \disc, \dist)\ p(\disc, \dist)$. 
$p(\inputs \mid \disc, \dist)$
is either generated (\eg in \cref{sec:linear}), or the empirical distribution of the subset of datapoints $\inputs$ with the corresponding underlying feature values (assuming access to these annotations, \eg in \cref{sec:imdb,sec:celeba}). $p(\disc, \dist)$ is varied across training conditions, as outlined below.

\textbf{Training conditions.} 
The upper-right quadrant in all subfigures of \cref{fig:setup-figure},
for which $p(\disc = 1, \dist = 1) = 1$,
acts as a hold-out set on which we can evaluate generalization to an unseen combination of attribute values. 
We produce multiple training conditions with the remaining three quadrants of data
by manipulating $p(\disc, \dist)$. All the analyses in this work compare model extrapolation to the held-out test quadrant across various training conditions.

To equalize the class base rates %
we balance all training conditions across 
the discriminant; \ie we enforce $p(\disc = 0) = p(\disc = 1) = 0.5$. We also fix the number of datapoints across all conditions at $N$;
With these constraints, we can control $p(\disc, \dist)$
via two degrees of freedom:
$\propzero = p(\dist = 1 \mid \disc = 0)$ (this implicitly fixes $ p(\dist = 0 \mid \disc = 0) = 1 - \propzero$ to balance the dataset); and
$\propone = p(\dist = 1 \mid \disc = 1)$.%
The three conditions in \cref{sec:expository}, as well as the held-out test set, correspond to particular settings of $\propzero$ and $\propone$ (shown in \cref{fig:setup-figure}, more in \cref{app:joint}).

\textbf{Measuring inductive bias.} 
We measure \textit{feature-level bias} as deviation from chance performance in the CC condition. 
\textit{Exemplar bias} is measured as the difference between performance in the \gls{pec} and \gls{zsc}----no difference indicates rule-based generalization, the magnitude of the difference measures exemplar bias.
Formally, 
for a given model family $\family$, 
let $\fzs$ denote the result of selecting a model from $\family$ 
by training in the zero-shot condition, and similarly $\fpe$ and $\fcc$. 
We define \gls{flb} and \gls{evr} as:
\begin{align}
  \operatorname{FLB}(\family) 
  &= \mathbb{E}[(A(y, \fcc(\inputs))] - 0.5~, \\
  \operatorname{EvR}(\family) 
  &= \mathbb{E}[ A(y, \fzs(\inputs))] -  \mathbb{E}[A(y, \fpe(\inputs))]~
\end{align}
where the expectation is taken with respect to the data distribution under the extrapolation region 
($p(\inputs, y \mid \propzero = 1, \propone = 1)$), and $A$ is the 0-1 accuracy. 
\gls{flb} takes values between -0.5 and 0.5 (indicating bias toward $\dist$ or $\disc$, respectively); 0 represents no feature bias. 
\gls{evr} takes values between 0 and 1 (indicating rule bias and exemplar bias, respectively). %

\textbf{Related formalisms and spurious correlation.} 
This binary formulation of discriminant and distractor features has previously been studied in the context of spurious correlation \citep{sagawa2020investigation}.
Rather than independently varying occupancy in the four quadrants, \citet{sagawa2020investigation} directly manipulate the (spurious) linear correlation between the distractor and the discriminant features ($p_{maj}$). 
In combinatorial feature spaces, a scalar spurious correlation insufficiently specifies the data distribution. 
The linear correlation coefficient $\corr$ between $\disc$ and $\dist$---henceforth \emph{spurious correlation}---can be written in terms of $\propzero$ and $\propone$
via
$\alpha = \frac{\propzero - \propone}{2}$ and 
$\beta = \frac{\propzero + \propone}{2}$ as 
\begin{align}
  \corr(\propzero, \propone) = \frac{\alpha}{\sqrt{\beta(1-\beta)}}~. \label{eq:spurious_corr}
\end{align}
Different combinations of $\propzero$ and $\propone$ give equal $\corr$ (see the contours in \cref{fig:interpolation-heatmap}, with markings for points along the equi-correlation contour from \gls{pe} ($\propzero = 0.5, \propone = 0.0$, $\corr = 0.58$));
 while nonetheless producing qualitatively different extrapolation behavior, as we demonstrate in later sections.
This indicates that sensitivity to spurious correlation insufficiently specifies extrapolation behavior. 
We argue for a formulation like ours---based on manipulating feature \textit{combinations}---that can tease apart distinct inductive biases at the level of what features a system finds easier to learn (\gls{flb}) as well as how to use these features to inform a decision boundary (\gls{evr}).

\bgroup

\setlength{\tabcolsep}{-4pt}
\renewcommand{\arraystretch}{1.5}

\begin{figure*}[t]
  \scriptsize

  \begin{subfigure}[b]{0.40\textwidth}

  \scalebox{0.85}{
  \begin{tabular}{
      M{25pt}
      M{80pt}
      M{80pt}
      M{80pt}
    }
    &
    \mbox{\hspace{16pt}cue conflict (CC)} &
    zero shot (ZS)  & 
    \mbox{\hspace{-12pt}partial exposure (PE)} 
    \\
    
    \normalsize\textbf{\acrshort{glm}}
    &
    \begin{tikzpicture}
      \begin{axis}[points-in-a-plane-axis, xmajorticks=false]
        \addplot graphics
        [xmin=-7, xmax=7, ymin=-7, ymax=7]
        {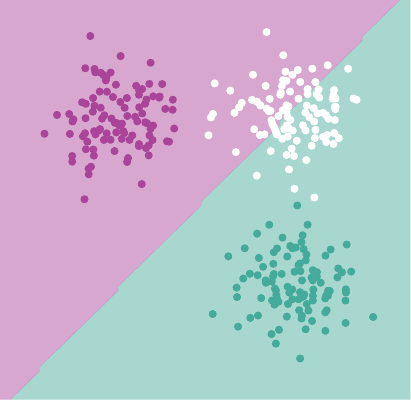};
      \end{axis}
    \end{tikzpicture}
    &
    \begin{tikzpicture}
      \begin{axis}[points-in-a-plane-axis, ticks=none]
        \addplot graphics 
        [xmin=-7, xmax=7, ymin=-7, ymax=7]
        {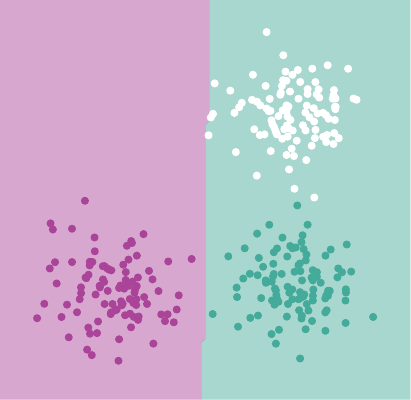};
      \end{axis}
    \end{tikzpicture}
    &
    \hspace{-14pt}
    \begin{tikzpicture}
      \begin{axis}[points-in-a-plane-axis, ticks=none]
        \addplot graphics 
        [xmin=-7, xmax=7, ymin=-7, ymax=7]
        {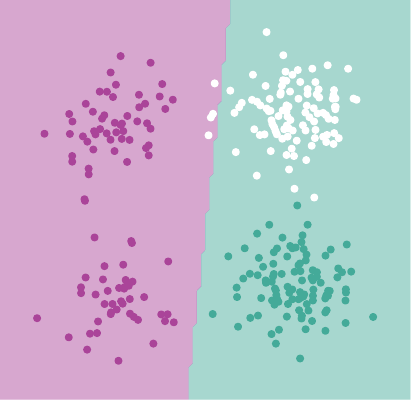};
      \end{axis}
    \end{tikzpicture} 
    \\

    \normalsize\textbf{\acrshort{gp}}
    &
    \begin{tikzpicture}
      \begin{axis}[points-in-a-plane-axis, xmajorticks=false]
        \addplot graphics
        [xmin=-7, xmax=7, ymin=-7, ymax=7]
        {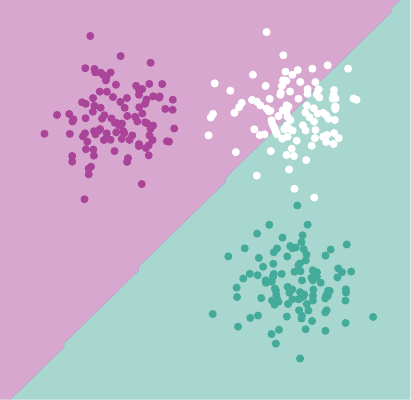};
      \end{axis}
    \end{tikzpicture}
    &
    \begin{tikzpicture}
      \begin{axis}[points-in-a-plane-axis, ticks=none]
        \addplot graphics 
        [xmin=-7, xmax=7, ymin=-7, ymax=7]
        {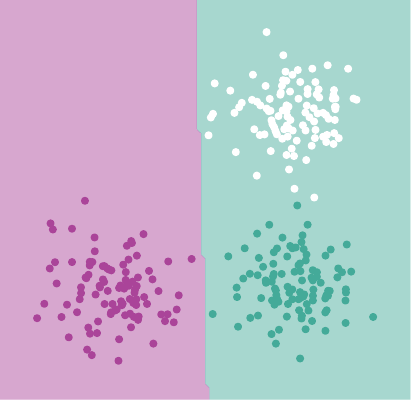};
      \end{axis}
    \end{tikzpicture}
    &
    \hspace{-14pt}
    \begin{tikzpicture}
      \begin{axis}[points-in-a-plane-axis, ticks=none]
        \addplot graphics 
        [xmin=-7, xmax=7, ymin=-7, ymax=7]
        {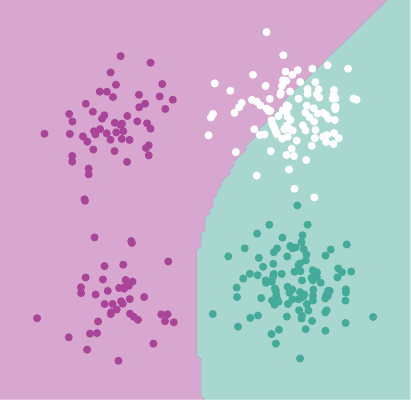};
      \end{axis}
    \end{tikzpicture} 
    \\

    \normalsize\textbf{\acrshort{nn}}
    &
    \begin{tikzpicture}
      \begin{axis}[points-in-a-plane-axis]
        \addplot graphics
        [xmin=-7, xmax=7, ymin=-7, ymax=7]
        {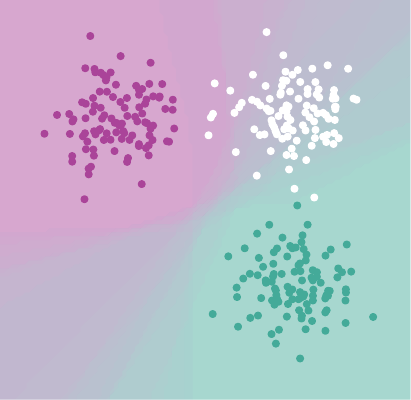};
      \end{axis}
    \end{tikzpicture}
    &
    \begin{tikzpicture}
      \begin{axis}[points-in-a-plane-axis, ymajorticks=false]
        \addplot graphics 
        [xmin=-7, xmax=7, ymin=-7, ymax=7]
        {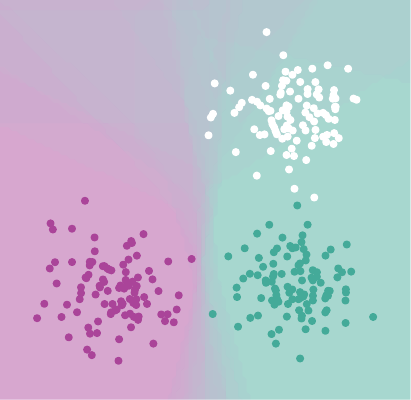};
      \end{axis}
    \end{tikzpicture}
    &
    \hspace{-14pt}
    \begin{tikzpicture}
      \begin{axis}[points-in-a-plane-axis, ymajorticks=false]
        \addplot graphics 
        [xmin=-7, xmax=7, ymin=-7, ymax=7]
        {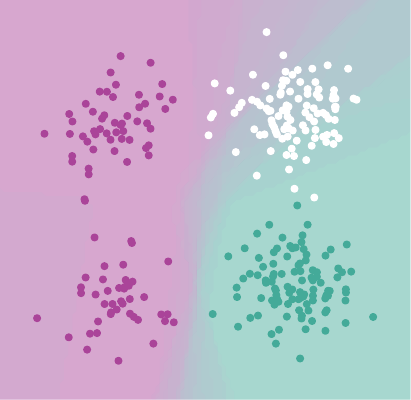};
      \end{axis}
    \end{tikzpicture} 
    \\

  \end{tabular}       
  }

    \caption{
      Decision boundaries averaged across 20 runs.
      Training datapoints are green or purple by label; test are white.
    }
    \label[figure]{fig:linear-decisions}
  \end{subfigure}
  \hfill
  \begin{subfigure}[b]{0.55\textwidth}
    \begin{tikzpicture}
      \begin{axis}[
          effect-bar,
          height=120pt,
          width=\linewidth,
          xlabel={model family $\family$},
          symbolic x coords={
            GLM:l1,
            GLM:l2,
            GLM:$\Phi$,
            \textbf{GLM:lin},
            null,
            GP:8.0,
            GP:0.5,
            \textbf{GP:fit},
            null,
            \textbf{NN:2h1d},
            NN:16h1d,
            NN:4h4d,
          },
        ]

        \addplot+[effect-bar-plot] 
        table[
          col sep=comma,
          header=true,
          x=model,
          y=zs-pe,
          y error=zs-pe-ci95,
        ] {figures/data/linear.csv};

      \end{axis}
    \end{tikzpicture}
    \caption{
      \textbf{\gls{evr} reflects exemplar-vs-rule propensity both within and across model families.}
      The \gls{evr} across model families, computed across 20 runs, error bars represent 95\% CIs. The GLMs are largely rule-based and show low \gls{evr}. Even within GLMs, sparsity regularization gives lower \gls{evr}. GPs are largely exemplar-based and show high \gls{evr}. Even within GPs, more `local' GPs with lower lengthscales have higher \gls{evr}. NNs lie in-between, with larger NNs giving higher \gls{evr}.
    }
    \label{fig:linear-results}
  \end{subfigure}

  \caption{
    \textbf{Simple 2-D classification} (\cref{sec:linear}) The specific model used in (a) are bolded in (b).
  } 
  \label{fig:linear}
\end{figure*}
\egroup
\section{2-D classification example}
\label{sec:linear}

To illustrate our framework in a simple statistical learning problem and quantitatively confirm the intuitions outlined in \cref{sec:expository}, we consider a two-dimensional classification problem. The feature dimensions are orthogonal bases in 2D space, and
we define the data generating procedure as
\begin{align}
    p(\inputs \mid \disc, \dist) &= \mathcal{N}(\mu, 1.0)~; \\
    \mu &= \alpha \times [2\disc -1, 2\dist -1]~, \notag
\end{align}
where, as specified in \cref{sec:setup},  $\disc, \dist, \in \{0, 1\}$, $p(\disc, \dist)$ is determined by the training condition. %
$\disc$ determines class labels, $\dist$ is a distractor, $\alpha$ is fixed at $3$, and $N = 300$ datapoints are in each class. The group with $\dist = \disc = 1$ is assigned the test set. 

\subsection{Model families and nomenclature.}
\label{sec:linearmodels}

\textbf{\Gls{nn}:} We train feedforward \gls{relu} classifiers with varying numbers of hidden layers and hidden units.
We use the scikit-learn implementation with default parameters, run 20 times for confidence intervals. 

\textbf{\Gls{glm}:} Parametric models allow us to formalize the feature-sparsity that characterizes rule-based learners. Linear logistic regression is sparse by definition (it has access to only linear features).
We generalize this model by expanding the feature space to include a nonlinear interaction $\Phi$ and examine L1 and L2 regularization in a \gls{glm} over this altered feature space.%

\textbf{\Gls{gp}:} 
Non-parametric kernel methods allow us to formalize exemplar-based generalization, where generalizations are made on the basis of feature-dense similarity to training data. 
We examine the performance of \gls{gp}s with \gls{rbf} kernels. We fit the kernel lengthscale using gradient descent on the log marginal likelihood of the data~\citep{rasmussen2003gaussian} (giving $5.2$) as well as vary it (adjusting ``locality'' in decision boundaries); GP:8.0 denotes a \gls{gp} with lengthscale value of 8.0.

We can implement explicit rule- and exemplar-based models in the synthetic setting since we know the features over which to build parametric or similarity-based models respectively, so we use it to validate our measures. In most application domains (including those in \cref{sec:imdb,sec:celeba}) feature learning is automated~\citep{hinton2006reducing}, making it difficult to specify the corresponding \gls{glm} or \gls{gp}.

\subsection{Comparing \gls{cc}, \gls{zs}, and \gls{pe}}
We consider one model from each class: \gls{nn} with 1 hidden layer of 2 units (NN:2h1d); linear \gls{glm} (GLM:lin); \gls{rbf} \gls{gp} with fitted lengthscale (GP:fit). The decision boundaries learned by these models are shown in \cref{fig:linear-decisions}. $\dist, \disc$ are equivalent by design, and permit no feature-level bias, so \gls{cc} is exactly at chance. This lets us focus on validating our novel protocol for measuring \gls{evr} without confounds. We generalize to cases with feature-level bias in later sections.
The \gls{glm}, sparse and therefore rule-based by definition, can only learn a linear boundary. It is therefore unaffected by the distractor dimension, showing no difference in extrapolation behavior between \gls{zs} and \gls{pe} (zero \gls{evr}). On the other hand, the \gls{gp} is exemplar-based by definition and displays a high \gls{evr}. The \gls{nn} shows an intermediate \gls{evr}, more rule-based than the purely-exemplar-based \gls{gp} but not entirely rule-based like the \gls{glm}.

\subsection{The influence  of model properties on \gls{evr}}
\label{sec:lin-model}

We first examine \gls{evr} in our control model classes (\gls{glm}s and \gls{gp}s) to validate that it tracks rule- vs exemplar-based extrapolation, followed by analyses of various \gls{nn}s.

\textbf{Regularized \gls{glm}s: \gls{evr} reduces with rule propensity.} A key property of rule propensity is sparsity in feature space. A linear \gls{glm} (GLM:lin) is sparse by definition, we examine a \gls{glm} on an expanded feature set so we can manipulate this sparsity. The additional feature $\Phi \propto \dist*\disc$ is the product of the observed features and normalizing by $\alpha$. We compute \gls{evr} for this \gls{glm} with different regularizers (regularization weight 1.0), shown in \cref{fig:linear-results}.

\Gls{glm} with no regularization (GLM:$\Phi$) displays a significant \gls{evr}. %
L2 regularization reduces it but L1 (which directly induces feature sparsity\footnote{Weight sparsity from L1-regularizer is equivalent to feature-sparsity only in special cases, including \gls{glm}.}) brings it to zero (or perfectly rule-based). This demonstrates that a low \gls{evr} tracks rule propensity via feature-level sparsity.

\textbf{Lengthscales in \gls{gp}s: \gls{evr} increases with exemplar propensity.} A sufficient condition for exemplar propensity is the locality of decision boundaries. We can directly manipulate this in a \gls{gp}s with its lengthscale. We evaluate \gls{evr} in \gls{gp}s with \gls{rbf} kernels of different lengthscales in \cref{fig:linear-results}. We find that the \gls{evr} is lowest with high lengthscales and grows as the lengthscale reduces, demonstrating that a high \gls{evr} tracks exemplar propensity via locality of decision boundaries.

\textbf{\Gls{nn}s: The necessary but insufficient role of expressivity.}
The results from \gls{glm}s and \gls{gp}s indicate that some ways to reduce expressivity (L1 regularization in \gls{glm}s and high lengthscale in \gls{rbf} \gls{gp}s) encourage rule propensity over exemplar propensity (thereby a lower \gls{evr}). We manipulate the most common variable in \gls{nn} expressivity---its size.

We increase the width of an \gls{nn} with fixed depth of 1 (\cref{fig:linear-results}) and find that the \gls{evr} increases. A deep \gls{nn} with the same number of units, however, exhibits comparable \gls{evr} to a wide network. Deeper networks with the same number of units are more expressive than wide ones \citep{raghu2017expressive}, indicating that excess expressivity, while necessary, is not the sole driver of \gls{evr}.

\subsection{\gls{evr} is distinct from sensitivity to spurious correlation}
A crucial difference between the \gls{zs} and the \gls{pe} conditions is that the \gls{pe} condition creates a (spurious) correlation $\corr = 0.58$ between $\dist$ and $\disc$. Is sensitivity to this spurious correlation ($\corr$) the sole the driver of the difference in performances between the \gls{pe} and \gls{zs} conditions, \ie of the \gls{evr}? We show that this is not the case; the \gls{evr} is measuring something distinct. As described in \cref{sec:setup}, there are multiple data-settings with the same $\corr$. We consider training conditions specified by other $\propzero, \propone$ that have the same $\corr$ as the \gls{pe} condition (dots along the solid contour in \cref{fig:interpolation-heatmap}). We find that performance on the extrapolation quadrant after training on these new data distributions is much higher (and closer to \gls{zs} performance) than when trained on the \gls{pe} condition---even though $\corr$ is exactly the same. This indicates that performance on the \gls{pe} condition (normalized by \gls{zs} performance to give the \gls{evr}) is uniquely indicative of something different from sensitivity to spurious correlations---it measures the inductive bias toward exemplar-vs-rule based extrapolation.

We can reduce $\corr$ in different ways by increasing $\propone$ or by reducing $\propzero$. We find that these are not equivalent and result in different extrapolation behaviors 
(\eg increasing $\propone$ gives more rule-based generalization than reducing $\propzero$; see results for the 2-D classification setting in \cref{sec:lin-interpolate} and for the vision domain in \cref{fig:celeba-interpolation}). 
This has implications for data manipulation methods (\eg subsampling or augmentation) that manipulate this $\corr$ to control extrapolation. This further supports that spurious correlation alone cannot explain extrapolation behavior, highlighting the importance of \gls{flb} and \gls{evr} that measure behavior under different feature \textit{combinations} in training.

\textbf{Conclusions.} 
\gls{evr} tracks exemplar- and rule-based extrapolation, as validated on interpretable models such as \gls{glm}s and \gls{gp}s. In particular, \gls{evr} decreases with reductions in expressivity mediated by regularization and lengthscale, and, in \gls{nn}s, also decreases with (some kinds of) expressivity. 
Finally, sensitivity to spurious correlation cannot explain the \gls{evr}.

\input{figures/05_imdb_wrapfig}
\section{IMDb text classification}
\label{sec:imdb}

In this section, we demonstrate our protocol on a standard text classification task: sentiment analysis on the \gls{imdb} dataset~\citep{maas2011learning}. 

\textbf{Selecting features.} The sentiment label (``positive'' or ``negative'') is the discriminant $\disc$. We manufacture an orthogonal distractor $\dist$ as the presence or absence of a word that occurs in roughly $50\%$ of the sentences in the dataset and does not occur more frequently for either positive or negative reviews. Some examples are ``film'' and ``you'': we use the word ``film'' (see \cref{fig:imdb}).

\textbf{Models.} We train a single-layer \gls{lstm}~\citep{hochreiter1997long} model of 20 hidden units on each condition and test on the held-out quadrant. We exclude models that do not reach $80\%$ validation accuracy.

\textbf{Feature-level bias.} 
The distractor $\dist$ is easier to learn than the discriminant $\disc$, as reflected in the \gls{cc} condition (19.7\%, $\operatorname{FLB} = -0.3$).%

\textbf{Exemplar bias.} 
We see good performance in \gls{zs} (84\%): Despite never having seen the word ``film,'' the system can generalize to reviews containing it. 
The performance in \gls{pe} drops significantly (30.1\%) giving a large \gls{evr} ($\operatorname{EvR} = 0.54$), indicating exemplar-based reasoning. 
As such, the exemplar-based tendency to utilize an additional unnecessary feature (\eg the presence of the word ``film'') hurts performance on the extrapolation quadrant. 
Performance in PE is higher than in CC, indicating that the system can learn to use the discriminant (\ie it is not purely relying on \gls{flb}).

\input{figures/06_celeba.tex}
\section{CelebA image classification}
\label{sec:celeba}

We now test our protocol on a standard classification task on a large-scale image dataset, \gls{celeba}~\citep{liu2015faceattributes}. Each image in this dataset is labeled with 40 binary attributes, each of which can be assigned discriminant or distractor. We examine \gls{flb} and \gls{evr} for standard models across different feature pairs, and discuss the practical implications of our findings.

\textbf{Selecting features.}
We select feature pairs that split the data roughly evenly and thus maximizing the number of training datapoints in each quadrant. We carry out our analyses across a range of feature pairs; an example is depicted in \cref{fig:celeba-quadrants}, and further details are in the Appendix.

\textbf{Models.}
We train \gls{resnet}~\citep{he2016deep} models of various depths ($\{10, 18, 34\}$) and widths ($\{2, 4, 8, 16, 32, 64\}$) on 6 different choices for feature pairs, with standard hyperparameters (see \cref{sec:datasets} for the complete feature set).
We limit our analyses to networks that achieve at least $75\%$ validation accuracy (on held-out samples from its own training distribution) to ensure that, despite differences in data variability across training conditions, all models learn a meaningful decision boundary.

\textbf{Feature-level bias.}
There is a wide range of \gls{flb} across feature pairs; \eg ``male'' is easier to learn than ``high cheekbones'' giving high \gls{flb}, and ``mouth open'' and ``wearing lipstick'' are equally difficult and give \gls{flb} of close to 0. \gls{flb} values for each feature pair were consistent across \gls{resnet} widths and depths.

\textbf{Exemplar-rule bias.} 
We observe good ZS performance: the models can generalize to new feature values outside the training support. We see a wide range of \gls{evr} across feature pairs, \cref{fig:celeba-effect}. Across all feature pairs, the \gls{evr} is non-negative: generalization in the PE condition is always worse (or not significantly better) than in the ZS condition. Further, we see a linear correlation between \gls{evr} and \gls{flb} in logit space across feature pairs. \gls{evr} therefore depends on how easy or hard the features are to learn. The key, however, is that this regression of the \gls{evr} onto \gls{flb} has a positive intercept: there is a positive \gls{evr} even for feature pairs with no \gls{flb}. That is, we see lower performance in PE compared to ZS (a nonzero \gls{evr}, exemplar propensity) even when \gls{flb} is controlled for.

We find no differences in \gls{evr} across \gls{resnet} widths and depths: \cref{fig:celeba-effect} plots \gls{evr} and \gls{flb} averaged over \gls{resnet} sizes.\footnote{We report width-and-depth-specific results in \cref{sec:width-depth}.}
One explanation is that the features in CelebA are complex; to learn these, we need reasonably high model expressivity, and differences in parameter count do not further modulate \gls{evr}. This is consistent with findings in \cref{sec:linear} where expressivity is necessary but not sufficient for increases in \gls{evr}: we see a jump in \gls{evr} going from NN:2h1d to NN:16h1d, but no further change going to the even more expressive NN:4h4d.

\textbf{Controlling spurious correlation.}
We replicate the findings in \cref{sec:linear}: the \gls{evr} cannot be explained by sensitivity to spurious correlation $\rho$. This is demonstrated in \cref{fig:celeba-interpolation}, where we substitute performance in the PE condition with performance in a different data condition ($\propzero = 0.825, \propzero = 0.25$) with the same $\rho = 0.58$ as in the PE condition. We find none of the effects discussed above, indicating that the PE condition is measuring something unique---exemplar-vs-rule propensity---which is not accounted for by sensitivity to spurious correlation. Further, \gls{evr} does not increase with model expressivity, unlike sensitivity to spurious correlation~\citep{sagawa2020investigation}.

\textbf{Practical implications of the \gls{evr}.} 
The nonzero \gls{evr} (\ie exemplar bias) reveals that  models are better at extrapolating zero-shot to a new feature value than when they have partial exposure to that feature value
\emph{even though the additional data need not change the learned decision boundary}. In particular, the training examples added in PE can be classified with the decision function from ZS without incurring additional training loss. A rule-based system recognizes this and bases its generalization on the minimal features that support the category boundary. However, an exemplar-based model changes its decision boundary in response to this additional data.%

PE-approximating data distributions ($\propzero \approx 0.5, \propone \approx 0.0$) occur naturally. 
For example, as \cite{sagawa2020investigation} observe, ``blond'' ``male''s are under-represented in \gls{celeba}. 
Consistent with the rest of our results, we find better classification for the extrapolation quadrant (blond males) if we discard data from an adjacent quadrant (blond non-males, or non-blond males) simulating the zero-shot condition, as opposed to the PE condition if such data is included: ResNet10, width 2, gives $\operatorname{ZS} = 75.12\pm 3.09\%$; $\operatorname{PE} = 60.22\pm 7.27\%$ for $\disc = $``male'' (discard blond non-males to get ZS) and $\operatorname{ZS} = 68.16\pm 3.34\%$; $\operatorname{PE} = 49.78\pm 3.76\%$ for $\disc = $``blond'' (discard non-blond males to get ZS). 

These results demonstrate the practical impact of understanding the exemplar-vs-rule bias in a model: an exemplar biased model (like the \gls{resnet} here) generalizes poorly in combinatorial settings, and can be made to generalize better by discarding an entire quadrant of data. Previous sub-sampling approaches \citep{sagawa2020investigation, haixiang2017learning} do not manipulate feature combinations and only manipulate spurious correlations. The aforementioned analyses (\cref{fig:interpolation-heatmap}) and results (\cref{fig:celeba-interpolation}) demonstrate that this underspecifies extrapolation behavior.

\section{Related work and future directions}
\label{sec:lit-rev}

\paragraph{Model design for systematic generalization.}
Rule-based generalization permits systematic extrapolation in combinatorial domains. 
This systematicity has been found lacking in neural networks~\citep{lake2017still, barrett2018measuring}, leading to renewed interest in hybrid symbolic--connectionist methods~\citep[\eg][]{garnelo2019reconciling}. 
However, works proposing new methods usually do not examine how feature co-occurrences modulate the systematicity of extrapolation. 
Using our protocol to examine exemplar- vs. rule-based generalization in these models is a promising future direction.

\textbf{Learning causal features.} 
Rule-based generalization, is equivalent to learning causal features under the assumption that the causal model is the simplest model that explains the data.
Recent work has investigated data settings that separate causal features from spurious ones~\citep[\eg][]{arjovsky2019invariant}.%
We showed that a model with exemplar propensity makes more rule-based extrapolations for certain training feature combinations (\ie \gls{zs} vs. \gls{pe}).
Investigating how feature coverage impacts causal generalization is a fruitful future direction.

\textbf{Similarity-based generalization and kernels.} 
We use similarity-based kernels (\eg \glsfirst{rbf}) to exemplify exemplar-based extrapolation. 
Recent work has interpreted neural networks as kernel regression~\citep{jacot2018neural}. %
Using a kernel framing to formalize the causes of exemplar bias is an exciting future direction.

\textbf{Data augmentation.} 
The \gls{evr} measure allows us to demonstrate that increased data variation in the form of feature coverage worsens systematic generalization. 
The negative effect of data variation on generalization has been documented for adversarial augmentations~\citep{raghunathan2020understanding}. 
We show that this can persist even when augmentation is not adversarial, rendering it generally relevant for the design of data augmentations.

\section{Conclusions}
\label{sec:conc}

Taking inspiration from---and going beyond---psychological studies, we design a behavioral protocol
to distinguish the effects of two 
inductive biases (feature-level bias and exemplar bias)
that is easily applicable to any classification domain.
This follows in a promising line of recent work that analyses and interprets deep learning systems based on their external behavior~\citep{ritter2017cognitive,dasgupta2019analyzing}. 
It complements other approaches that follow in the neuroscience tradition of analyzing internal representations~\citep{zeiler2014visualizing, karpathy2015visualizing}
or make approximations of these internal workings to support theoretical results~\citep{jacot2018neural,allen2019convergence}. 
The behavioral approach has the advantage that it makes no assumptions about the model, allowing comparisons across systems that differ in design.

Both rule- and exemplar-based extrapolation are valuable depending on domain, underscoring the importance of diagnosing feature-level bias and exemplar bias.
Moreover, studying this trade-off allows us to demonstrate an important phenomenon:
We find that more feature coverage (as in \gls{pe} compared to \gls{zs}) hurts generalization for exemplar-based models. %
This has implications for methods that manipulate data distributions to improve performance 
(\eg data subsampling~\citep{haixiang2017learning}, 
data augmentation~\citep{perez2017effectiveness}, 
and contrastive learning~\citep{chen2020simple}).
Since an exemplar-based model tends to acquire spurious associations, 
our measures have the potential to be useful as diagnostics in application settings where the goal is to control model behavior on non-representative factors (\citep[\eg][]{mitchell2019model}).

A limitation of the present work is that we do not provide a conclusive answer as to what properties of a model family influence both feature-level bias and exemplar bias.
A broader study of these factors and theoretical work formalizing this effect are exciting avenues for future work.

\section*{Acknowledgements and Funding Sources}
We thank our anonymous reviewers for their feedback.
This work was supported by 
ONR Grant \#N00014-18-1-2873
and the DARPA L2M program.

\bibliographystyle{front-matter/icml-style/icml2022}
\bgroup
\bibliography{refs}
\egroup

\clearpage
\onecolumn
\appendix

\section{Additional formalizations}

\subsection{Generalizing the framework from two binary attributes to many categorical attributes}
In the most general terms, we consider a setting in which each observation $\mathbf{x} \in \mathcal{X}$ is underlied by $n$ categorical variables $z_1, \dots, z_n \in \{0, \dots, C\}$ with $C \in \mathbb{Z}_+$, henceforth \emph{attributes}
whose concatenation $\attribute = (z_1, \dots, z_n)$ determines the observable input $\inputs$
via some mapping $g : \mathbb{Z}_{0+}^n \to \mathcal{X}$.
We consider the binary classification task of fitting a model $\hat{f} : \mathcal{X} \to \{0, 1\}$ from a given model family $\mathcal{F}$
to predict a binary label for each input.
A subset of the attributes in $\attribute$, without loss of generality $(z_0, \dots, z_i)$, is taken to define the decision boundary, while the remaining attributes, $z_{i+1}, \dots, z_{n}$, are assumed to not be independently predictive of the true classification $y \in \{0, 1\}$.
We therefore denote the \emph{discriminant}, $\disc = (z_0, \dots, z_i)$,
and the \emph{distractor} $\dist = (z_{i+1}, \dots, z_n)$.
For simplicity, we assume that the attributes are binary (\ie $C=2$ and $z_i \in \{0, 1\}, \forall i$), and
that the discriminant attributes must be jointly active for the classification to change from the null class $y=0$
(\ie $y = 1 \iff \disc = \mathbf{1}$);
the latter simplification allows us to redefine
$\disc = z_0 \wedge \cdots \wedge z_i$ and
$\dist = z_{i+1} \wedge \cdots \wedge z_n$, which is equivalent to the earlier discussion of the illustrative two-attribute case.

\subsection{Training conditions expressed in terms of the joint distribution}
\label{app:joint}
\begin{figure*}[t]
  \centering
  \footnotesize

  \begin{tabular}{
      M{71pt}
      M{66pt}
      M{66pt}
      M{66pt}
      M{66pt}
    }

    \toprule

    &
    \multicolumn{3}{c}{training condition} & 
    \\
    \cmidrule(rl){2-4}
    feature space & 
    \gls{cc} & 
    \gls{zs} & 
    \gls{pe} &
    extrapolation
    \\

    \scriptsize ${\color{muted-purple}\propzero=0.5}$, ${\color{muted-teal}\propone=0.5}$
    & \scriptsize ${\color{muted-purple}\propzero=1.0}$, ${\color{muted-teal}\propone=0.0}$
    & \scriptsize ${\color{muted-purple}\propzero=0.0}$, ${\color{muted-teal}\propone=0.0}$
    & \scriptsize ${\color{muted-purple}\propzero=0.5}$, ${\color{muted-teal}\propone=0.0}$
    & \scriptsize ${\color{muted-purple}\propzero=1.0}$, ${\color{muted-teal}\propone=1.0}$
    \\\midrule

    \hspace{-10pt}
    \begin{tikzpicture}
      \scriptsize
      \fill[muted-purple!15] (0,0) -- (0,2.2) -- (1.1,2.2) -- (1.1,0);
      \fill[muted-teal!15] (1.1,0) -- (1.1,2.2) -- (2.2,2.2) -- (2.2,0);

      \draw[->,thick] (0, 0) -- (2.2, 0) node[midway,below,align=center] {\textbf{discriminant} (color)};
      \draw[->,thick] (0, 0) -- (0, 2.2) node[midway,left,rotate=90,anchor=south,align=center] {\textbf{distractor} (shape)};

      \node[diamond-stimulus,purple-stimulus] (off-off) at (0.6,0.6) {};
      \node[diamond-stimulus,green-stimulus] (on-off) at (1.6,0.6) {};
      \node[circle-stimulus,purple-stimulus] (off-on) at (0.6,1.6) {};
      \node[circle-stimulus,green-stimulus] (on-on) at (1.6,1.6) {};
    \end{tikzpicture}
    &
    \begin{tikzpicture}[baseline={(0,-0.35)}]
      \scriptsize
      \fill[muted-purple!15] (0,0) -- (0,2.2) -- (1.1,2.2) -- (1.1,0);
      \fill[muted-teal!15] (1.1,0) -- (1.1,2.2) -- (2.2,2.2) -- (2.2,0);

      \draw[->,thick] (0, 0) -- (2.2, 0);
      \draw[->,thick] (0, 0) -- (0, 2.2);

      \node[diamond-stimulus,green-stimulus] (on-off) at (1.6,0.6) {};
      \node[circle-stimulus,purple-stimulus] (off-on) at (0.6,1.6) {};
    \end{tikzpicture}
    &
    \begin{tikzpicture}[baseline={(0,-0.35)}]
      \scriptsize
      \fill[muted-purple!15] (0,0) -- (0,2.2) -- (1.1,2.2) -- (1.1,0);
      \fill[muted-teal!15] (1.1,0) -- (1.1,2.2) -- (2.2,2.2) -- (2.2,0);

      \draw[->,thick] (0, 0) -- (2.2, 0);
      \draw[->,thick] (0, 0) -- (0, 2.2);

      \node[diamond-stimulus,purple-stimulus] (off-off) at (0.6,0.6) {};
      \node[diamond-stimulus,green-stimulus] (on-off) at (1.6,0.6) {};
    \end{tikzpicture}
    &
    \begin{tikzpicture}[baseline={(0,-0.35)}]
      \scriptsize
      \fill[muted-purple!15] (0,0) -- (0,2.2) -- (1.1,2.2) -- (1.1,0);
      \fill[muted-teal!15] (1.1,0) -- (1.1,2.2) -- (2.2,2.2) -- (2.2,0);

      \draw[->,thick] (0, 0) -- (2.2, 0);
      \draw[->,thick] (0, 0) -- (0, 2.2);

      \node[diamond-stimulus,purple-stimulus] (off-off) at (0.6,0.6) {};
      \node[diamond-stimulus,green-stimulus] (on-off) at (1.6,0.6) {};
      \node[circle-stimulus,purple-stimulus] (off-on) at (0.6,1.6) {};
    \end{tikzpicture}
    & 
    \begin{tikzpicture}
      \scriptsize
      \fill[muted-purple!15] (0,0) -- (0,2.2) -- (1.1,2.2) -- (1.1,0);
      \fill[muted-teal!15] (1.1,0) -- (1.1,2.2) -- (2.2,2.2) -- (2.2,0);

      \draw[->,thick] (0, 0) -- (2.2, 0);
      \draw[->,thick] (0, 0) -- (0, 2.2);

      \node[circle-stimulus,green-stimulus] (on-on) at (1.6,1.6) {};
    \end{tikzpicture}

     \\

     \begin{tikzpicture}
       \hspace{-15pt}
       \begin{axis}[
           points-in-a-plane-axis,
           xlabel={\textbf{discriminant} ($x_1 > 0$)},
           x label style={yshift=10pt,align=center,font=\scriptsize},
           ylabel={\textbf{distractor} ($x_2 > 0$)},
           y label style={yshift=-20pt,align=center,font=\scriptsize},
           clip=false,
       ]

       \node[anchor=west,font=\scriptsize] (x1) at (axis cs:6.7, -7) {\bm{$x_1$}};
       \node[anchor=south,font=\scriptsize] (x2) at (axis cs:-7, 6.7) {\bm{$x_2$}};

       \addplot[muted-purple,points-in-a-plane,samples=50] ({invgaussx(rnd,rnd,-3,1)}, {invgaussy(rnd,rnd,-3,1)});
       \addplot[muted-purple,points-in-a-plane,samples=50] ({invgaussx(rnd,rnd,-3,1)}, {invgaussy(rnd,rnd,3,1)});
       \addplot[muted-teal,points-in-a-plane,samples=50] ({invgaussx(rnd,rnd,3,1)}, {invgaussy(rnd,rnd,-3,1)});
       \addplot[muted-teal,points-in-a-plane,samples=50] ({invgaussx(rnd,rnd,3,1)}, {invgaussy(rnd,rnd,3,1)});

       \end{axis}
     \end{tikzpicture}
     &
     \begin{tikzpicture}
       \hspace{-12.5pt}
       \begin{axis}[
           points-in-a-plane-axis,
       ]
       \addplot[muted-purple,points-in-a-plane,samples=100] ({invgaussx(rnd,rnd,-2.5,1)}, {invgaussy(rnd,rnd,2.5,1)});
       \addplot[muted-teal,points-in-a-plane,samples=100] ({invgaussx(rnd,rnd,2.5,1)}, {invgaussy(rnd,rnd,-2.5,1)});

       \end{axis}
     \end{tikzpicture}
     &
     \begin{tikzpicture}
       \hspace{-12.5pt}
       \begin{axis}[
           points-in-a-plane-axis,
       ]
       \addplot[muted-purple,points-in-a-plane,samples=100] ({invgaussx(rnd,rnd,-2.5,1)}, {invgaussy(rnd,rnd,-2.5,1)});
       \addplot[muted-teal,points-in-a-plane,samples=100] ({invgaussx(rnd,rnd,2.5,1)}, {invgaussy(rnd,rnd,-2.5,1)});

       \end{axis}
     \end{tikzpicture}
     &
     \begin{tikzpicture}
       \hspace{-12.5pt}
       \begin{axis}[
           points-in-a-plane-axis,
       ]
       \addplot[muted-purple,points-in-a-plane,samples=50] ({invgaussx(rnd,rnd,-2.5,1)}, {invgaussy(rnd,rnd,-2.5,1)});
       \addplot[muted-purple,points-in-a-plane,samples=50] ({invgaussx(rnd,rnd,-2.5,1)}, {invgaussy(rnd,rnd,2.5,1)});
       \addplot[muted-teal,points-in-a-plane,samples=100] ({invgaussx(rnd,rnd,2.5,1)}, {invgaussy(rnd,rnd,-2.5,1)});

       \end{axis}
     \end{tikzpicture}
     &
     \begin{tikzpicture}
       \hspace{-12.5pt}
       \begin{axis}[
           points-in-a-plane-axis,
       ]
       \addplot[muted-teal,points-in-a-plane,samples=100] ({invgaussx(rnd,rnd,2.5,1)}, {invgaussy(rnd,rnd,2.5,1)});

       \end{axis}
     \end{tikzpicture}
  \end{tabular}       
                     
  \label[figure]{fig:full-setup-figure}
  \caption{
    We expand on \cref{fig:setup-figure} from the main text by including a realization of the abstract training conditions in the simple 2D points-in-a-plane setting.
    (\textbf{Top}) 
    \textbf{Formalizing the illustrative experiment:} The experiment from \cref{fig:expository} expressed in terms of the formalism in \cref{sec:setup} with $\dist = $ color and $\disc = $ shape. Background colors indicate true category boundary.
    (\textbf{Bottom}) The conditions realized via a binarization of continuous feature values.
     Here, the discriminant is binarized as $x_1 > 0$ 
     and the distractor as $x_2 > 0$; this setting is further investigated in \cref{sec:linear}.
     Color here depicts the label but is not part of the input.
  }

\end{figure*}
We express the training conditions displayed in \cref{fig:setup-figure} and realized in 
Figure 6
in terms of the joint distribution instead of the parameters $\propzero$, $\propone$.

\begin{enumerate}[leftmargin=20pt]
  \item The \gls{ccc} the upper left and lower right quadrants in Figure 6
    and defines the distribution of attributes as

    \begin{center}
    \begin{tabular}{
        >{\color{muted-purple}}r
        >{\hspace{-9.5pt}}>{\color{muted-purple}}l
        |
        >{\color{muted-green}}r
        >{\hspace{-9.5pt}}>{\color{muted-green}}l
      }
      $p_\text{cc}(\disc = 0, \dist = 1)$ &= $0.5$ &
      $p_\text{cc}(\disc = 1, \dist = 1)$ &= $0$ \\\midrule
      $p_\text{cc}(\disc = 0, \dist = 0)$ &= $0$ &
      $p_\text{cc}(\disc = 1, \dist = 0)$ &= $0.5$~.
    \end{tabular}
    \end{center}

  \item The \gls{zsc} populates the bottom left and right quadrants in Figure 6
    and defines the distribution of attributes as

    \begin{center}
    \begin{tabular}{
        >{\color{muted-purple}}r
        >{\hspace{-9.5pt}}>{\color{muted-purple}}l
        |
        >{\color{muted-green}}r
        >{\hspace{-9.5pt}}>{\color{muted-green}}l
      }
      $p_\text{zs}(\disc = 0, \dist = 1)$ &= $0$ &
      $p_\text{zs}(\disc = 1, \dist = 1)$ &= $0$ \\\midrule
      $p_\text{zs}(\disc = 0, \dist = 0)$ &= $0.5$ &
      $p_\text{zs}(\disc = 1, \dist = 0)$ &= $0.5$~.
    \end{tabular}
    \end{center}

  \item The \gls{pec} populates all quadrants but the upper right in Figure 6
    and defines the distribution of attributes as

    \begin{center}
    \begin{tabular}{
        >{\color{muted-purple}}r
        >{\hspace{-9.5pt}}>{\color{muted-purple}}l
        |
        >{\color{muted-green}}r
        >{\hspace{-9.5pt}}>{\color{muted-green}}l
      }
      $p_\text{pe}(\disc = 0, \dist = 1)$ &= $0.25$ &
      $p_\text{pe}(\disc = 1, \dist = 1)$ &= $0$ \\\midrule
      $p_\text{pe}(\disc = 0, \dist = 0)$ &= $0.25$ &
      $p_\text{pe}(\disc = 1, \dist = 0)$ &= $0.5$~.
    \end{tabular}
    \end{center}
\end{enumerate}

\section{CelebA results for specific model sizes}
\label{sec:width-depth}

\begin{figure}[ht]

  \begin{subfigure}[b]{0.3\textwidth}
  \includegraphics[width=\textwidth]{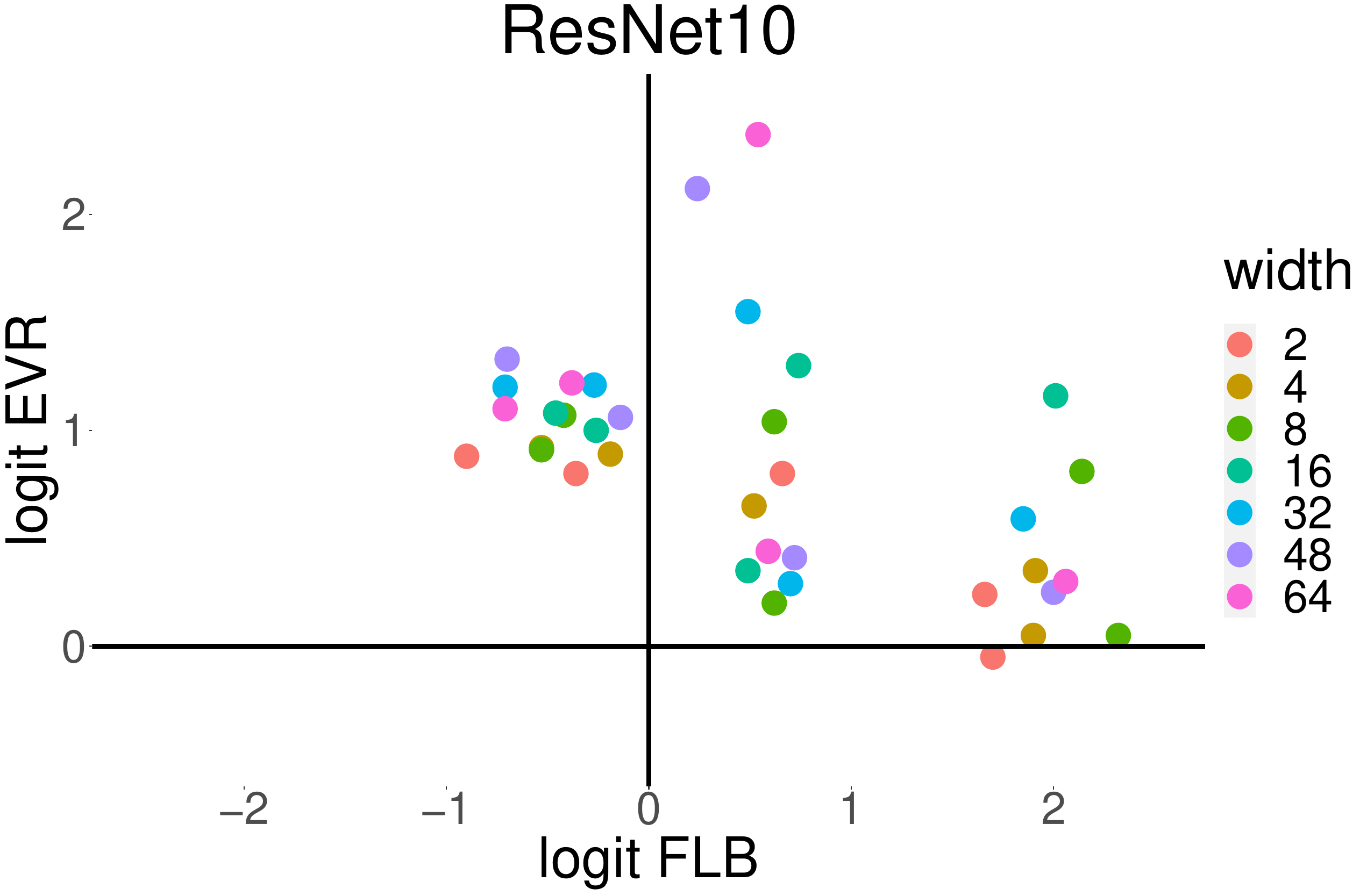}
  \end{subfigure}
  \hfill
  \begin{subfigure}[b]{0.3\textwidth}
  \includegraphics[width=\textwidth]{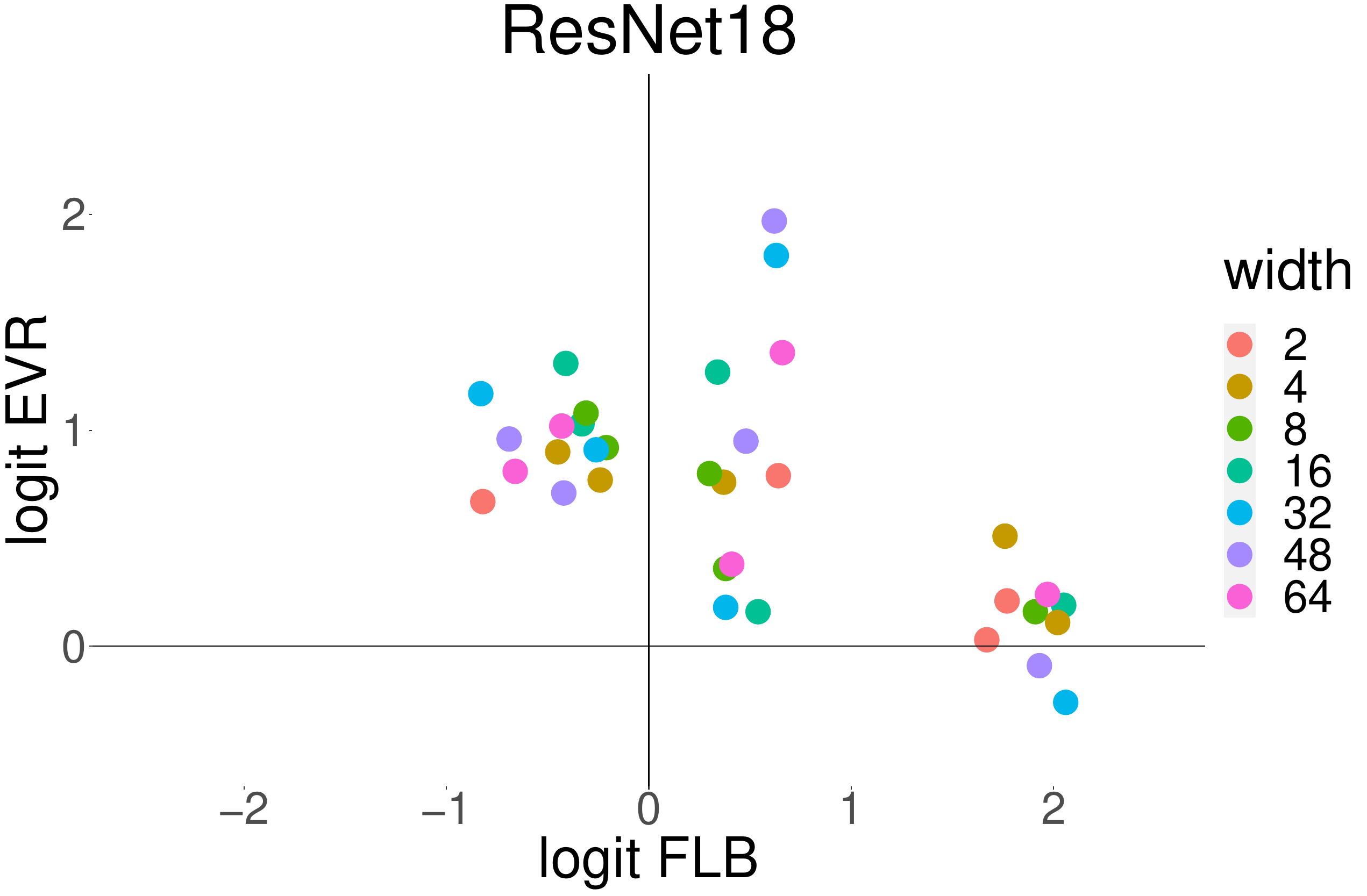}
  \end{subfigure}
  \hfill
  \begin{subfigure}[b]{0.3\textwidth}
  \includegraphics[width=\textwidth]{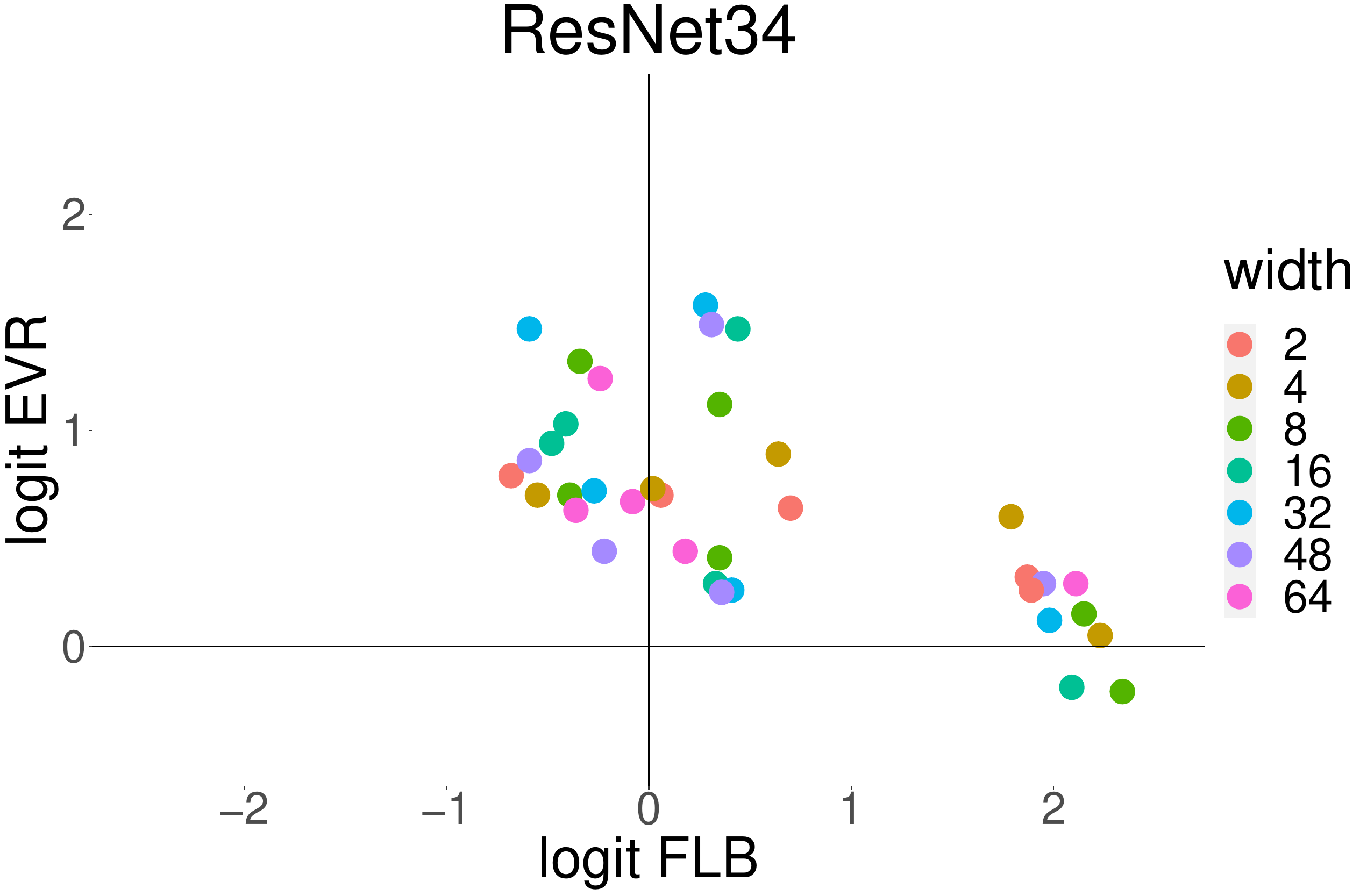}
  \end{subfigure}
  
  \caption{CelebA \gls{evr} and \gls{flb} across feature pairs, averaged across ~30 runs, split by depth and width of ResNet.}
 \vspace{-15pt}
  \label{fig:celeba-splits}
\end{figure}

We include model-specific results, split by ResNet depth and width, in \cref{fig:celeba-splits}.
We find no systematic relationship between \gls{evr} and depth or width.

\section{Spurious correlation underdetermines feature distributions} 
\label{sec:supp-interpolation}

The partial-exposure condition ($\propzero = 0.5$, $\propone = 0.0$) in \cref{sec:expository} results in a spurious correlation between the discriminant $\disc$ and the distractor $\dist$ ($\rho = 0.58$). To examine behavior in a wider range of data settings, we vary $\propzero$ and $\propone$ as described in \cref{sec:setup}, thereby also changing the degree of spurious correlation.

\textbf{I. Interpolation towards zero shot.} 
\label{sec:zero-shot-interpolation}
We interpolate $\propzero$ from $0.5$ towards $0.0$, keeping $\propone = 0.0$. This moves us closer to $\propzero = \propone = 0.0$, where we have no exposure to $\disc = 1$ in training. Intuitively, we are reducing the exposure to the new distractor feature value from the \gls{pec}.

\textbf{II. Interpolation to full exposure.} 
\label{sec:full-exposure-interpolation}
We interpolate $\propone$ from $0.0$ towards $0.5$, keeping $\propzero = 0.5$. This moves us closer to $\propzero = \propone = 0.5$, where we have equal exposure to all quadrants in training. Here, rather than reducing the exposure to the new distractor feature value, we are equalizing the exposure to it across the discriminant dimension.

\textbf{III. Interpolation with matched correlation.} We report results on this in \cref{sec:linear,sec:celeba}.
As also depicted in \cref{fig:interpolation-heatmap}, we generate training conditions by changing $\propzero$ and $\propone$ such that we follow a $\corr$-contour away from the partial-exposure condition ($\propzero = 0.5, \propone = 0.0$, $\corr = 0.58$): solid contour in \cref{fig:interpolation-heatmap}. We also match the spurious correlation across the two interpolations in \cref{sec:supp-interpolation}A and B: \cref{fig:interpolation} shows these additional $\corr$-contours as dashed lines.

These different interpolations are depicted in \cref{fig:interpolation-heatmap-supp} with different shape/colors.

\subsection{Generating interpolation points}
\label{sec:interpolation-procedure}

We generate points along all three interpolation lines: 
from partial exposure towards zero shot~(\labelcref{sec:supp-interpolation}I);
from partial exposure towards full exposure~(\labelcref{sec:supp-interpolation}II); 
and the equi-correlation line originating from partial exposure~(\labelcref{sec:supp-interpolation}III).
The interpolating points along each line are selected to balance spurious correlation and feature exposure.
In particular, we follow the following procedure:
\begin{enumerate}[leftmargin=20pt]
    \item We choose a point that interpolates towards full exposure. We do this by choosing a value of $\propone$ between $0.0$ and $0.5$, $\propfe$. This gives a data setting, along with a corresponding spurious correlation, $\corr$, computed via \cref{eq:spurious_corr}:
      \begin{align*}
        \propzero &= 0.5~; &
        \propone &= \propfe~; &
        \corr &= \corr\left(0.5, \propfe\right)~.
      \end{align*}

    \item We generate a corresponding point that interpolates towards zero shot. Given the data setting above, we set $\propone = 0.0$ and compute the $\propzero$ to produce the same $\corr$ as the full-exposure interpolations in Step 1. This gives the data setting:
      \begin{align*}
        \propzero &= \propzs~; &
        \propone &= 0.0~; &
        \corr &= \corr\left(\propzs, 0.0\right)
        =\corr\left(0.5, \propfe\right)~.
      \end{align*}

    \item Finally, we also derive the equi-correlation interpolation from the full-exposure interpolation as follows. We retain $\propone$ from the full-exposure condition, but recompute the $\propzero$ such that the correlation $\corr$ matches the spurious correlation of the pure \\gls{pec} ($\corr = 0.58$). This gives an additional data setting:
      \begin{align*}
        \propzero &= \propeq~; &
        \propone &= \propfe; &
        \corr &= \corr\left(0.5, 0.0\right)
        = 0.58~.
      \end{align*}
    
\end{enumerate}

Note that, despite there being three different interpolation lines, the specific interpolants we use are constrained along a single degree of freedom---choosing $\propfe$ (Step 1). The data settings for zero shot (Step 2) and equi-correlation (Step 3) are derived from this value.

\subsection{Specific interpolation values used}
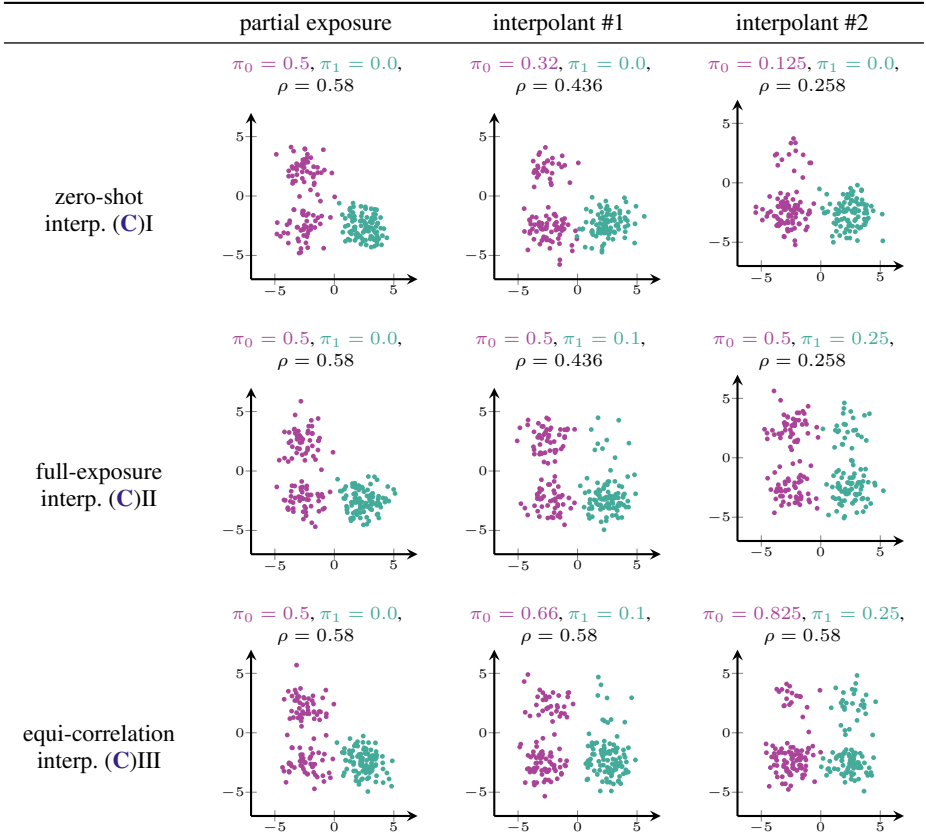
\begin{figure*}[t]
  \footnotesize
  \centering

  \begin{tabular}{
      M{60pt}
      M{80pt}
      M{80pt}
      M{80pt}
    }

    \toprule

    &
    \gls{pe} &
    interpolant \#1 &
    interpolant \#2 
    \\\midrule

    &
    \scriptsize ${\color{muted-purple}\propzero=0.5}$, ${\color{muted-teal}\propone=0.0}$, $\rho=0.58$
    & \scriptsize ${\color{muted-purple}\propzero=0.32}$, ${\color{muted-teal}\propone=0.0}$, $\rho=0.436$
    & \scriptsize ${\color{muted-purple}\propzero=0.125}$, ${\color{muted-teal}\propone=0.0}$, $\rho=0.258$
    \\

    zero-shot interp.~\labelcref{sec:supp-interpolation}I
     &
     \begin{tikzpicture}
       \begin{axis}[
           points-in-a-plane-axis,
       ]
       \addplot[muted-purple,points-in-a-plane,samples=50] ({invgaussx(rnd,rnd,-2.5,1)}, {invgaussy(rnd,rnd,-2.5,1)});
       \addplot[muted-purple,points-in-a-plane,samples=50] ({invgaussx(rnd,rnd,-2.5,1)}, {invgaussy(rnd,rnd,2.5,1)});
       \addplot[muted-teal,points-in-a-plane,samples=100] ({invgaussx(rnd,rnd,2.5,1)}, {invgaussy(rnd,rnd,-2.5,1)});

       \end{axis}
     \end{tikzpicture}
    &
     \begin{tikzpicture}
       \begin{axis}[
           points-in-a-plane-axis,
       ]

       \addplot[muted-purple,points-in-a-plane,samples=68] ({invgaussx(rnd,rnd,-2.5,1)}, {invgaussy(rnd,rnd,-2.5,1)});
       \addplot[muted-purple,points-in-a-plane,samples=32] ({invgaussx(rnd,rnd,-2.5,1)}, {invgaussy(rnd,rnd,2.5,1)});
       \addplot[muted-teal,points-in-a-plane,samples=100] ({invgaussx(rnd,rnd,2.5,1)}, {invgaussy(rnd,rnd,-2.5,1)});

       \end{axis}
     \end{tikzpicture}
     &
     \begin{tikzpicture}
       \begin{axis}[
           points-in-a-plane-axis,
       ]
       \addplot[muted-purple,points-in-a-plane,samples=87] ({invgaussx(rnd,rnd,-2.5,1)}, {invgaussy(rnd,rnd,-2.5,1)});
       \addplot[muted-purple,points-in-a-plane,samples=13] ({invgaussx(rnd,rnd,-2.5,1)}, {invgaussy(rnd,rnd,2.5,1)});
       \addplot[muted-teal,points-in-a-plane,samples=100] ({invgaussx(rnd,rnd,2.5,1)}, {invgaussy(rnd,rnd,-2.5,1)});

       \end{axis}
     \end{tikzpicture}

    \\

    &
    \scriptsize ${\color{muted-purple}\propzero=0.5}$, ${\color{muted-teal}\propone=0.0}$, $\rho=0.58$
    & \scriptsize ${\color{muted-purple}\propzero=0.5}$, ${\color{muted-teal}\propone=0.1}$, $\rho=0.436$
    & \scriptsize ${\color{muted-purple}\propzero=0.5}$, ${\color{muted-teal}\propone=0.25}$, $\rho=0.258$
    \\

    full-exposure interp.~\labelcref{sec:supp-interpolation}II
    &
     \begin{tikzpicture}
       \begin{axis}[
           points-in-a-plane-axis,
       ]
       \addplot[muted-purple,points-in-a-plane,samples=50] ({invgaussx(rnd,rnd,-2.5,1)}, {invgaussy(rnd,rnd,-2.5,1)});
       \addplot[muted-purple,points-in-a-plane,samples=50] ({invgaussx(rnd,rnd,-2.5,1)}, {invgaussy(rnd,rnd,2.5,1)});
       \addplot[muted-teal,points-in-a-plane,samples=100] ({invgaussx(rnd,rnd,2.5,1)}, {invgaussy(rnd,rnd,-2.5,1)});

       \end{axis}
     \end{tikzpicture}
     &
     \begin{tikzpicture}
       \begin{axis}[
           points-in-a-plane-axis,
       ]
       \addplot[muted-purple,points-in-a-plane,samples=50] ({invgaussx(rnd,rnd,-2.5,1)}, {invgaussy(rnd,rnd,-2.5,1)});
       \addplot[muted-purple,points-in-a-plane,samples=50] ({invgaussx(rnd,rnd,-2.5,1)}, {invgaussy(rnd,rnd,2.5,1)});
       \addplot[muted-teal,points-in-a-plane,samples=90] ({invgaussx(rnd,rnd,2.5,1)}, {invgaussy(rnd,rnd,-2.5,1)});
       \addplot[muted-teal,points-in-a-plane,samples=10] ({invgaussx(rnd,rnd,2.5,1)}, {invgaussy(rnd,rnd,2.5,1)});

       \end{axis}
     \end{tikzpicture}
     &
     \begin{tikzpicture}
       \begin{axis}[
           points-in-a-plane-axis,
       ]
       \addplot[muted-purple,points-in-a-plane,samples=50] ({invgaussx(rnd,rnd,-2.5,1)}, {invgaussy(rnd,rnd,-2.5,1)});
       \addplot[muted-purple,points-in-a-plane,samples=50] ({invgaussx(rnd,rnd,-2.5,1)}, {invgaussy(rnd,rnd,2.5,1)});
       \addplot[muted-teal,points-in-a-plane,samples=75] ({invgaussx(rnd,rnd,2.5,1)}, {invgaussy(rnd,rnd,-2.5,1)});
       \addplot[muted-teal,points-in-a-plane,samples=25] ({invgaussx(rnd,rnd,2.5,1)}, {invgaussy(rnd,rnd,2.5,1)});

       \end{axis}
     \end{tikzpicture}

    \\

    &
    \scriptsize ${\color{muted-purple}\propzero=0.5}$, ${\color{muted-teal}\propone=0.0}$, $\rho=0.58$
    & \scriptsize ${\color{muted-purple}\propzero=0.66}$, ${\color{muted-teal}\propone=0.1}$, $\rho=0.58$
    & \scriptsize ${\color{muted-purple}\propzero=0.825}$, ${\color{muted-teal}\propone=0.25}$, $\rho=0.58$
    \\

    equi-correlation interp.~\labelcref{sec:supp-interpolation}III
    &
     \begin{tikzpicture}
       \begin{axis}[
           points-in-a-plane-axis,
       ]

       \addplot[muted-purple,points-in-a-plane,samples=50] ({invgaussx(rnd,rnd,-2.5,1)}, {invgaussy(rnd,rnd,-2.5,1)});
       \addplot[muted-purple,points-in-a-plane,samples=50] ({invgaussx(rnd,rnd,-2.5,1)}, {invgaussy(rnd,rnd,2.5,1)});
       \addplot[muted-teal,points-in-a-plane,samples=100] ({invgaussx(rnd,rnd,2.5,1)}, {invgaussy(rnd,rnd,-2.5,1)});

       \end{axis}
     \end{tikzpicture}
     &
     \begin{tikzpicture}
       \begin{axis}[
           points-in-a-plane-axis,
       ]
       \addplot[muted-purple,points-in-a-plane,samples=66] ({invgaussx(rnd,rnd,-2.5,1)}, {invgaussy(rnd,rnd,-2.5,1)});
       \addplot[muted-purple,points-in-a-plane,samples=34] ({invgaussx(rnd,rnd,-2.5,1)}, {invgaussy(rnd,rnd,2.5,1)});
       \addplot[muted-teal,points-in-a-plane,samples=90] ({invgaussx(rnd,rnd,2.5,1)}, {invgaussy(rnd,rnd,-2.5,1)});
       \addplot[muted-teal,points-in-a-plane,samples=10] ({invgaussx(rnd,rnd,2.5,1)}, {invgaussy(rnd,rnd,2.5,1)});

       \end{axis}
     \end{tikzpicture}
     &
     \begin{tikzpicture}
       \begin{axis}[
           points-in-a-plane-axis,
       ]
       \addplot[muted-purple,points-in-a-plane,samples=82] ({invgaussx(rnd,rnd,-2.5,1)}, {invgaussy(rnd,rnd,-2.5,1)});
       \addplot[muted-purple,points-in-a-plane,samples=18] ({invgaussx(rnd,rnd,-2.5,1)}, {invgaussy(rnd,rnd,2.5,1)});
       \addplot[muted-teal,points-in-a-plane,samples=75] ({invgaussx(rnd,rnd,2.5,1)}, {invgaussy(rnd,rnd,-2.5,1)});
       \addplot[muted-teal,points-in-a-plane,samples=25] ({invgaussx(rnd,rnd,2.5,1)}, {invgaussy(rnd,rnd,2.5,1)});

       \end{axis}
     \end{tikzpicture}
  \end{tabular}       
                     
  \label[figure]{fig:celeba-interpolants}
  \caption{We visualize several of the interpolants used for the interpolation analyses.}

\end{figure*}

For all data settings, we generate points along the interpolation lines using the procedure in \cref{sec:interpolation-procedure}.

For the simple 2D classification setting, we examine two interpolants. In this simple domain, we keep the interpolation distances small, since we expect changes in extrapolation behavior even from small changes.

\begin{center}
\begin{tabular}{l ccc ccc}
  \toprule
  & \multicolumn {3}{c}{interpolant 1} & \multicolumn {3}{c}{interpolant 2} \\\cmidrule(rl){2-4} \cmidrule(rl){5-7}
                                 & $\propzero$ & $\propone$ & $\corr$ & $\propzero$ & $\propone$ & $\corr$ \\
  \midrule
  interpolation to zero shot~(\labelcref{sec:supp-interpolation}I)             & $0.481$ & $0.0$  & $0.563$ & $0.32$  & $0.0$ & $0.436$ \\
  interpolation to full exposure~(\labelcref{sec:supp-interpolation}II)         & $0.5$   & $0.01$ & $0.563$ & $0.5$   & $0.1$ & $0.436$ \\
  equi-correlation interpolation~(\labelcref{sec:supp-interpolation}III) & $0.519$ & $0.01$ & $0.58$  & $0.661$ & $0.1$ & $0.58$  \\
  \bottomrule
\end{tabular}
\end{center}

For CelebA, we increase the interpolation distance to reflect the wider range of natural data distributions among feature pairs. The data these interpolation values generate is visualized as the equivalent points-in-a-plane setting in 
Figure 7.

\begin{center}
\begin{tabular}{l ccc ccc}
  \toprule
  & \multicolumn {3}{c}{interpolant 1} & \multicolumn {3}{c}{interpolant 2} \\\cmidrule(rl){2-4} \cmidrule(rl){5-7}
                                 & $\propzero$ & $\propone$ & $\corr$ & $\propzero$ & $\propone$ & $\corr$ \\
  \midrule
  interpolation to zero shot~(\labelcref{sec:supp-interpolation}I)             & $0.32$  & $0.0$  & $0.436$ & $0.125$  & $0.0$  & $0.258$ \\
  interpolation to full exposure~(\labelcref{sec:supp-interpolation}II)         & $0.5$   & $0.1$  & $0.436$ & $0.5$    & $0.25$ & $0.258$ \\
  equi-correlation interpolation~(\labelcref{sec:supp-interpolation}III) & $0.66$  & $0.1$  & $0.58$  & $0.825$  & $0.25$ & $0.58$  \\
  \bottomrule
\end{tabular}
\end{center}

\begin{figure*}[t]
  \begin{subfigure}[b]{0.48\textwidth}
  \centering
    \begin{tikzpicture}[heatmap-spy]
      \begin{axis}[
          heatmap-axis,
          height=120pt,
          width=120pt,
        ]

        \addplot graphics
        [xmin=-0.04, xmax=1.04, ymin=-0.04, ymax=1.04]
        {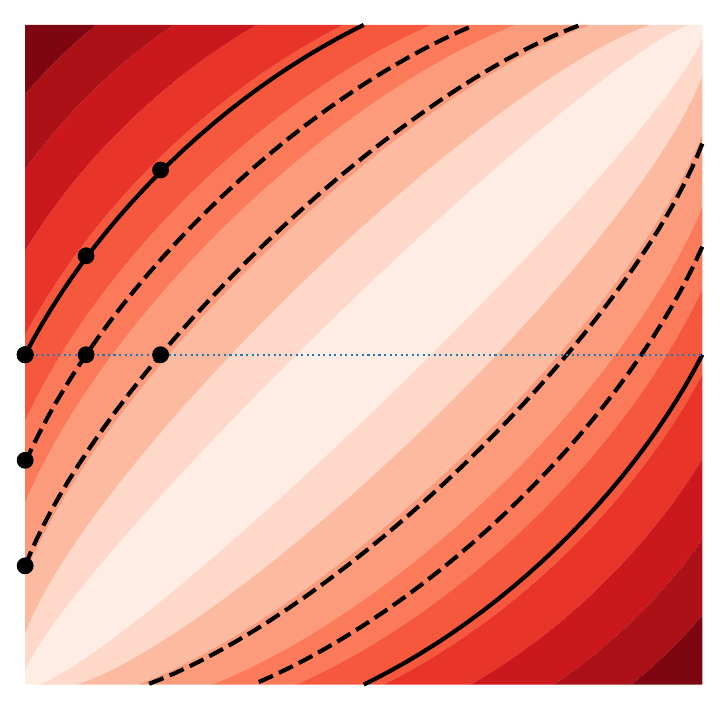};

        \node[pe-mark] at (axis cs: 0.0, 0.5) {};

        \node[zs-interp-mark] at (axis cs: 0.0, 0.34) {};
        \node[zs-interp-mark] at (axis cs: 0.0, 0.18) {};
        \node[fe-interp-mark] at (axis cs: 0.09, 0.5) {};
        \node[fe-interp-mark] at (axis cs: 0.2, 0.5) {};
        \node[corr-interp-mark] at (axis cs: 0.088, 0.65) {};
        \node[corr-interp-mark] at (axis cs: 0.20, 0.78) {};

        \coordinate (fe-interp) at (axis cs:0.1, 0.5);
        \coordinate (zs-interp) at (axis cs:0, 0.34);
        \coordinate (corr-interp) at (axis cs:0.09, 0.65);

      \end{axis}

      \spy[lens={rotate=90, scale=0.8}, width=30pt, vibrant-teal, very thick] on (zs-interp) in node (zs-spy) at (3.2, 2.25);
      \spy[lens={rotate=0, scale=0.8}, width=25pt, vibrant-blue, very thick] on (fe-interp) in node (fe-spy) at (3.2, 1.25);
      \spy[lens={rotate=-54, scale=0.8}, width=30pt, vibrant-cyan, very thick] on (corr-interp) in node (corr-spy) at (3.2, 0.25);

      \node[anchor=west,align=center] at (3.7, 2.25) {\tiny zero-shot interp.~\labelcref{sec:supp-interpolation}A};
      \node[anchor=west,align=center] at (3.7, 1.25) {\tiny full-exposure interp.~\labelcref{sec:supp-interpolation}B};
      \node[anchor=west,align=center] at (3.7, 0.25) {\tiny equi-correlation interp.~\labelcref{sec:supp-interpolation}C};

    \end{tikzpicture}
    \caption{A heatmap of the spurious correlation (\cref{eq:spurious_corr}), showing different interpolations.
    The \gls{pec} is identified with a star outline; points along the three 
    interpolation line are identified with filled shapes.
    }
    \label{fig:interpolation-heatmap-supp}
  \end{subfigure}
  \hfill
  \begin{subfigure}[b]{0.22\textwidth}

    \begin{tikzpicture}
      \begin{axis}[
          interpolation-axis,
        ]

        \addplot[interp-plot] table [col sep=space, x=dist, y=pee, y error=ci95] {
          rho0 rho1 dist pee ci95
          0.50 0.0 0.0  0.36966667 0.053843826
          0.48 0.00 0.02 0.33060000 0.049656405
          0.32 0.00 0.18 0.30996667 0.055412473
        };

        \addplot[interp-plot] table [col sep=space, x=dist, y=pee, y error=ci95] {
          rho0 rho1 dist pee ci95
          0.50 0.0 0.0  0.36966667 0.053843826
          0.50 0.01 0.01 0.08253333 0.017150132
          0.50 0.10 0.1 -0.01303333 0.001259950
        };

        \addplot[interp-plot] table [col sep=space, x=dist, y=pee, y error=ci95] {
          rho0 rho1 dist pee ci95
          0.50 0.0 0.0  0.36966667 0.053843826
          0.52 0.01 0.02236   0.14063333 0.024030974
          0.66 0.10 0.1886796 -0.00660000 0.001817273
        };

        \node[pe-mark] at (axis cs: 0.0, 0.36966667) {};

        \node[zs-interp-mark] at (axis cs: 0.02, 0.33060000) {};
        \node[zs-interp-mark] at (axis cs: 0.18, 0.30996667) {};

        \node[fe-interp-mark] at (axis cs: 0.01, 0.08253333) {};
        \node[fe-interp-mark] at (axis cs: 0.1, -0.01303333) {};

        \node[corr-interp-mark] at (axis cs: 0.02236, 0.14063333) {};
        \node[corr-interp-mark] at (axis cs: 0.1886796, -0.0066000) {};

      \end{axis}
    \end{tikzpicture}
    \caption{
      Interpolations for NN:16h1d on 2D classification of points-in-a-plane.
    }
    \label{fig:interpolation-linear}
  \end{subfigure}
  \hfill
  \begin{subfigure}[b]{0.2\textwidth}
    \begin{tikzpicture}
      \begin{axis}[
          interpolation-axis,
          ylabel={},
          xmax=0.5,
        ]

        \addplot[interp-plot] table [col sep=space, x=dist, y=pee, y error=ci95] {
          rho0 rho1 dist pee ci95
          0.5 0.0 0.0 0.3468000000000001 0.1707
          0.32 0.0 0.18 0.08250000000000002 0.0688
          0.125 0.0 0.375 0.05020000000000013 0.08960000000000001

        };

        \addplot[interp-plot] table [col sep=space, x=dist, y=pee, y error=ci95] {
          rho0 rho1 dist pee ci95
          0.5 0.0 0.0 0.3468000000000001 0.1707
          0.5 0.1 0.1 0.10150000000000003 0.08439999999999999
          0.5 0.25 0.25 0.07469999999999999 0.0454
        };

        \addplot[interp-plot] table [col sep=space, x=dist, y=pee, y error=ci95] {
          rho0 rho1 dist pee ci95
          0.5 0.0 0.0 0.3468000000000001 0.1707
          0.66 0.1 0.1886796226411321   0.10250000000000004 0.0506
          0.825 0.25 0.4100304866714181 0.08979999999999999 0.048499999999999995
        };

        \node[pe-mark] at (axis cs: 0.0, 0.3468000000000001) {};

        \node[zs-interp-mark] at (axis cs: 0.18, 0.08250000000000002) {};
        \node[zs-interp-mark] at (axis cs: 0.375, 0.05020000000000013) {};

        \node[fe-interp-mark] at (axis cs: 0.1, 0.10150000000000003) {};
        \node[fe-interp-mark] at (axis cs: 0.25, 0.07469999999999999) {};

        \node[corr-interp-mark] at (axis cs: 0.1886796226411321,  0.10250000000000004) {};
        \node[corr-interp-mark] at (axis cs: 0.4100304866714181,  0.08979999999999999) {};

      \end{axis}
    \end{tikzpicture}
    \caption{
      Interpolations, ResNet-18-8, \acrshort{celeba} (``wearing lipstick'', ``mouth open'')
    }
    \label{fig:celeba-interpolation-supp}
  \end{subfigure}

  \caption{Interpolations away from the PE: changes in extrapolation behavior under data distribution with the same spurious correlation as in PE, as well as different ways to change spurious correlation.
 }
 \vspace{-15pt}
  \label{fig:interpolation}
\end{figure*}
\subsection{Interpolation analyses}

\subsubsection{In the 2-D classification example}
\label{sec:lin-interpolate}

In the simple setting from \cref{sec:linear}, we vary $\propzero$, $\propone$ for an \gls{nn} model (NN:16h1d, the \gls{nn} with lowest \gls{evr} level overall). Results are in \cref{fig:interpolation-linear} and discussed below.

\gls{evr} $\neq$ \textbf{sensitivity to spurious correlation.} As also discussed in the main text, along the equi-correlation interpolation line, the ``effective \gls{evr}'' drops drastically (\ie the learner generalizes in more rule-based manner) despite no change in spurious correlation. 

\textbf{Implications for controlling extrapolation.} Despite both having the same $\corr$, interpolating towards full-exposure increases the \gls{evr} more than towards zero-shot. This further supports that spurious correlation cannot fully characterize extrapolation behavior. This shows that different \textit{ways} to reduce $\corr$ have different effects on extrapolation, and has important implications for data manipulation methods (\eg subsampling or augmentation) that aim to directly control this $\corr$.

\subsubsection{In CelebA}
We see the same effects as in the linear setting: as also discussed in the main text, we see a much smaller gap to the ZS condition despite no change in spurious correlation. We don't find clear effects distinguishing different ways to reduce spurious correlation (interpolation to zero shot~(\labelcref{sec:supp-interpolation}I) and interpolation to full exposure~(\labelcref{sec:supp-interpolation}II)).

\section{Reproducibility details}
\label{sec:reprodicibility}

We use the criteria from \cite{MLRepro}
omitting the non-applicable \emph{theory} component.

\subsection{Models \& Algorithms}

\paragraph{An analysis of the complexity (time, space, sample size) of any algorithm.}
The algorithms we employ (LBFGS and stochastic gradient descent on convex and nonconvex problems) are standard, 
and so we refer the reader to other references to determine their complexities.

\subsection{Datasets}
\label{sec:datasets}
\paragraph{The relevant statistics, such as number of examples.}

\begin{center}
\begin{tabular}{llll}
\toprule
  & \texttt{2D} 
  & \texttt{IMDb} 
  & \texttt{CelebA}\footnotemark \\
\midrule
  dataset size (train) & 75 
  & 21,215 
  & 4,000 to 40,000 \\
  dataset size (valid) & 75 
  & 21,027 
  & 8,000 \\
  dataset size (test) & 75 
  & 13,995 
  & 20,000 \\
  input space & $\mathbb{R}^2$ 
  & $\mathbb{R}^{400}$ 
  & $\mathbb{R}^{178 \times 218 \times 3}$ \\
\bottomrule
\end{tabular}
\end{center}

\footnotetext{The numbers for \acrshort{celeba} are approximate because there are  deviations in the availability of images across attribute combinations.}

\paragraph{The details of train / validation / test splits.}
We do not use a validation set for the simple 2D classification setting
and IMDb datasets, 
but hold out examples for a test set.
For CelebA, we follow the authors' division of images in train, validation and test splits.

\paragraph{An explanation of any data that were excluded, and all pre-processing step.}
As described in the main text, we subsample data to balance attributes within each training condition.
For the \acrshort{celeba} domain,
we use the following feature pairs to produce the results in \cref{sec:celeba}:

\begin{center}
\begin{tabular}{ll}
  \toprule 
  \textbf{discriminant} & \textbf{distractor} \\
  \midrule 
  mouth open & male \\
  wearing lipstick & mouth open \\
  male & mouth open \\
  male & high cheekbones \\
  male & blond hair \\
  male & arched eyebrows \\
  \bottomrule
\end{tabular}
\end{center}

\paragraph{A link to a downloadable version of the dataset or simulation environment.}
We provide code to generate the points-in-a-plane dataset of \cref{sec:linear} in our code repository, whose link is below in \cref{sec:code}.
For \cref{sec:imdb} and \cref{sec:celeba}, we used publicly available datasets:
\gls{imdb}~\citep{maas2011learning} is available at 
\url{https://ai.stanford.edu/~amaas/data/sentiment/};
\gls{celeba}~\citep{liu2015faceattributes} at 
\url{https://mmlab.ie.cuhk.edu.hk/projects/CelebA.html}.

\paragraph{For new data collected, a complete description of the data collection process, such as
instructions to annotators and methods for quality control.}
We do not collect any new data.

\subsection{Code}
\label{sec:code}
See the code repository located at 
\url{https://github.com/eringrant/icml-2022-rules-vs-exemplars}.

\subsection{Experimental results}
\paragraph{The range of hyper-parameters considered, method to select the best hyper-parameter
configuration, and specification of all hyper-parameters used to generate results.}
We use default hyperparameter settings whenever possible; all hyperparameter settings can be 
found in the `configs` folder of our code repository, whose link is below in \cref{sec:code}.

\paragraph{The exact number of training and evaluation runs.}
For the points-in-a-plane and \acrshort{imdb} settings, we use 20 random seeds,
which randomize the model weight initialization.
For the \acrshort{celeba} domain, we run 30 seeds for each model configuration, and discard runs that achieve below 75\% accuracy on validation set images that belong to the data conditions (quadrants) observed during training.

\paragraph{A clear definition of the specific measure or statistics used to report results.}
We report accuracy as a performance metric on each of the four quadrants depicted in \cref{fig:setup-figure} as well interpolating data settings. We additionally report measures that are a the performance difference between data settings.

\paragraph{A description of results with central tendency (\eg mean) \& variation (\eg error bars).}
See the main text for a description of results.
We include a 95\% confidence interval on all reported measures.

\paragraph{The average runtime for each result, or estimated energy cost.}
For the points-in-a-plane, \acrshort{imdb} and \acrshort{celeba} datasets, the average runtime (time to train and evaluate a single model) is 5, 10 and 30 minutes, respectively.

\paragraph{A description of the computing infrastructure used.}
We run experiments serially on an NVIDIA P100 GPU.

\end{document}